\documentclass[letterpaper, 10 pt, conference]{ieeeconf}  

\IEEEoverridecommandlockouts                        
\title{\LARGE \bf
    $\mathcal{D(R, O)}$ Grasp: A Unified Representation of Robot and Object Interaction for Cross-Embodiment Dexterous Grasping
}

\author{Zhenyu Wei$^{1,2*}$, Zhixuan Xu$^{1*}$, Jingxiang Guo$^{1}$, Yiwen Hou$^{1}$, \\ Chongkai Gao$^{1}$, Zhehao Cai$^{1}$, Jiayu Luo$^{1}$, Lin Shao$^{1}$\textsuperscript{\textdagger}%
\thanks{* denotes equal contribution}%
\thanks{\textdagger \ denotes the corresponding author}%
\thanks{$^{1}$Zhenyu Wei (internship), Zhixuan Xu, Jingxiang Guo, Yiwen Hou, Chongkai Gao, Zhehao Cai, Jiayu Luo, and Lin Shao are with the Department of Computer Science, National University of Singapore. 
        {\tt\small zhixuanxu@u.nus.edu, linshao@nus.edu.sg}}%
\thanks{$^{2}$Zhenyu Wei is with the School of Electronic Information and Electrical Engineering, Shanghai Jiao Tong University, and the Zhiyuan College, Shanghai Jiao Tong University.
        {\tt\small Zhenyu\_Wei@sjtu.edu.cn}}%
}

\usepackage{bbm}
\usepackage{etoolbox}
\usepackage{multicol}
\usepackage{cite}
\usepackage[bookmarks=true]{hyperref}
\usepackage{graphics} 
\usepackage{times} 
\usepackage{amsmath} 
\usepackage{amssymb}  
\usepackage{svg}
\usepackage{mathrsfs}
\usepackage{amsfonts}
\usepackage{wrapfig}
\usepackage{amsmath}

\DeclareMathOperator*{\argmin}{arg\,min}
\usepackage{kantlipsum}
\usepackage{subcaption}
\usepackage{marvosym}

\usepackage{graphicx}
\usepackage{float}
\usepackage{ctable}
\usepackage{cuted}
\usepackage{colortbl}
\usepackage{multirow}
\usepackage{wrapfig}
\usepackage[misc,geometry]{ifsym}
\usepackage{algorithm}
\usepackage{algpseudocode}
\usepackage{xcolor}   
\usepackage{pifont}    

\usepackage{bm}
\usepackage{caption}
\usepackage{setspace}
\let\labelindent\relax
\usepackage{enumitem}
\usepackage{threeparttable}
\newcommand{\TODO}[1][]{\textcolor{red}{\bf [TODO]}}
\setlist[enumerate,1]{itemsep=3pt}

\definecolor{formalgreen}{rgb}{0.1, 0.7, 0.1}  
\definecolor{formalred}{rgb}{0.9, 0.2, 0.2}  

\newcommand{\cmark}{\textcolor{formalgreen}{\checkmark}}  
\newcommand{\xmark}{\textcolor{formalred}{\ding{55}}}           

\begin{document}

\maketitle
\thispagestyle{empty}
\pagestyle{empty}

\begin{abstract}
Dexterous grasping is a fundamental yet challenging skill in robotic manipulation, requiring precise interaction between robotic hands and objects. In this paper, we present $\mathcal{D(R, O)}$ Grasp, a novel framework that models the interaction between the robotic hand in its grasping pose and the object, enabling broad generalization across various robot hands and object geometries. Our model takes the robot hand's description and object point cloud as inputs and efficiently predicts kinematically valid and stable grasps, demonstrating strong adaptability to diverse robot embodiments and object geometries. Extensive experiments conducted in both simulated and real-world environments validate the effectiveness of our approach, with significant improvements in success rate, grasp diversity, and inference speed across multiple robotic hands. Our method achieves an average success rate of 87.53\% in simulation in less than one second, tested across three different dexterous robotic hands. In real-world experiments using the LeapHand, the method also demonstrates an average success rate of 89\%. $\mathcal{D(R,O)}$ Grasp provides a robust solution for dexterous grasping in complex and varied environments. The code, appendix, and videos are available on our project website at~\href{https://nus-lins-lab.github.io/drograspweb/}{https://nus-lins-lab.github.io/drograspweb/}.
\end{abstract}
\section{Introduction}
Dexterous grasping is crucial in robotics as the first step in executing complex manipulation tasks. However, quickly obtaining a high-quality and diverse set of grasps remains challenging for dexterous robotic hands due to their high degrees of freedom and the complexities involved in achieving stable, precise grasps. Researchers have developed several optimization-based methods to address this challenge ~\cite{roa2015grasp, dfc, chen2024springgrasp, patel2024getzero,haldar2023teach}. Some of these methods, however, often focus on fingertip point contact, rely on complete object geometry, and require significant computational time to optimize. As a result, data-driven grasp generation methods have gained attention. These methods aim to solve the grasping problem using learning-based techniques. We can broadly categorize them into two types: those that utilize robot-centric representations, such as wrist poses and joint values~\cite{xu2023unidexgrasp, xu2023fast, wan2023unidexgrasp++}, and those that rely on object-centric representations, such as contact points~\cite{shao2020unigrasp,attarian2023geometry,li2021end} or contact maps~\cite{li2023gendexgrasp,xu2024manifoundation,morrison2018closing,varley2015generating}.

Robot-centric representations (e.g., joint values), as used in methods like UniDexGrasp++~\cite{wan2023unidexgrasp++}, directly map observation to control commands, enabling fast inference. However, these methods often exhibit low sample efficiency. Furthermore, these approaches struggle to generalize across different robotic embodiments, as the learned mappings are specific to the training data and do not quickly adapt to new robot designs or geometries. Object-centric representations (e.g., key points, contact points, affordances) effectively capture the geometry and contacts of objects, allowing for generalization across different shapes and robots, as demonstrated by methods like UniGrasp~\cite{shao2020unigrasp} and GenDexGrasp~\cite{li2023gendexgrasp}. However, these methods are often less efficient as they typically require an additional optimization step—such as solving fingertip inverse kinematics (IK) or fitting the predicted contact maps under penetration-free and joint limit constraints to translate the object-centric representation into actionable robot commands. This optimization process is time-consuming due to its complexity and nonconvexity~\cite{wu2022learning}.

\begin{figure}[t] \centering
    \includegraphics[width=0.9 \linewidth]{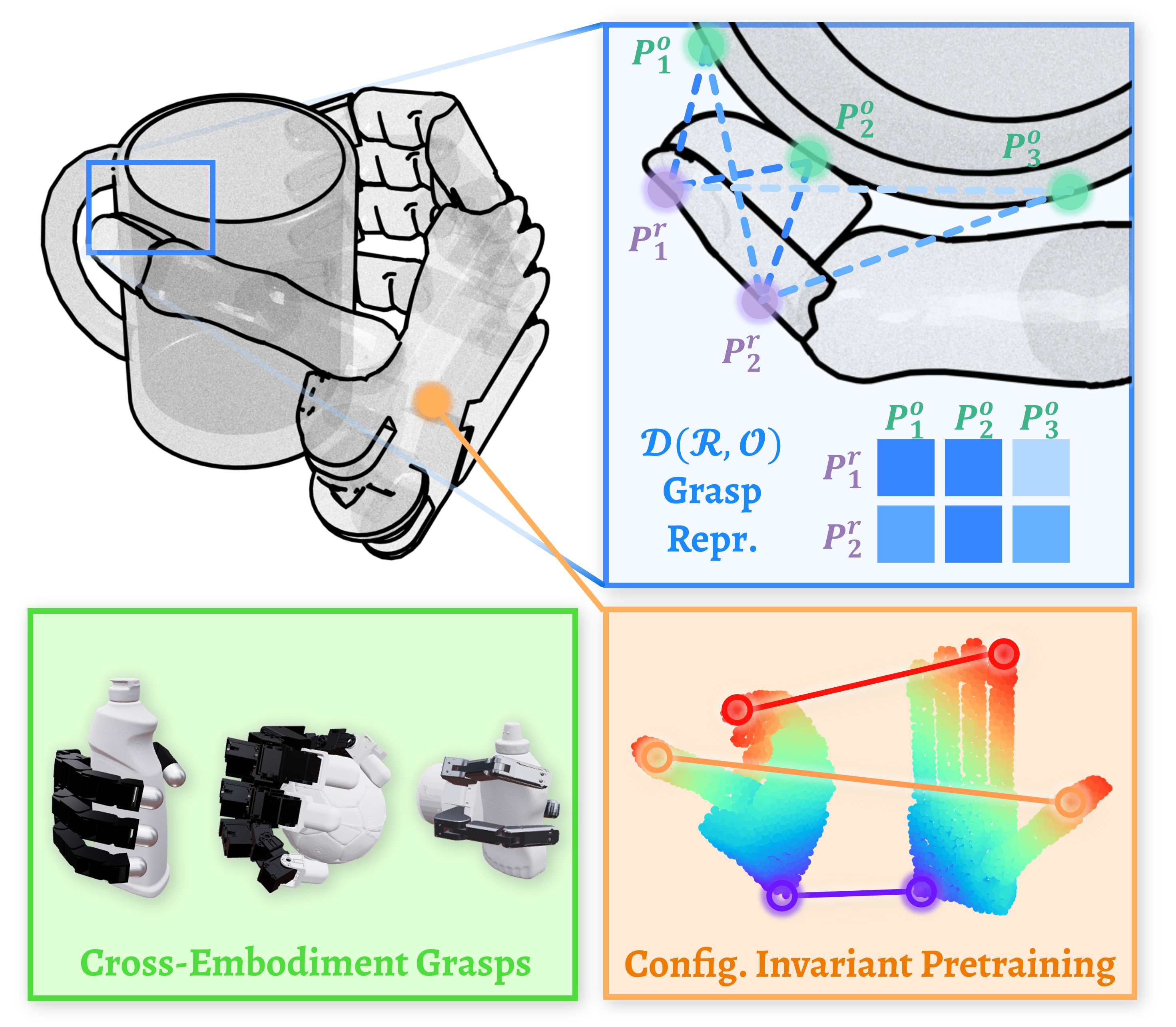}
    \vspace{-1pt}
    \caption{We propose our model that utilizes configuration-invariant pretraining, predicts $\mathcal{D(R,O)}$ representation, and obtains grasps for cross-embodiment from point cloud input.}
    \label{fig:figure1}
    \vspace{-15pt}
\end{figure}

\begin{table*}[htbp]
    \resizebox{\textwidth}{!}{
        \begin{tabular}{@{}ccccccccc@{}}
            \toprule
            & \begin{tabular}[c]{@{}c@{}}Grasp\\ Representation\end{tabular} & Method Type & \begin{tabular}[c]{@{}c@{}}Cross\\ Embodiment\end{tabular} & \begin{tabular}[c]{@{}c@{}}Inference\\ Speed\end{tabular} & \begin{tabular}[c]{@{}c@{}}Sample\\ Efficiency\end{tabular} & \begin{tabular}[c]{@{}c@{}}Partial Object\\ Point Cloud\end{tabular} & \begin{tabular}[c]{@{}c@{}}Full-hand Contact\\ (not only fingertips)\end{tabular} & \begin{tabular}[c]{@{}c@{}}Optional Grasp\\ Preference Interface\end{tabular} \\
            \midrule
            DFC \cite{dfc} & Joint Values & Robot-centric & \cmark & \xmark \xmark & - & \xmark & \cmark & \xmark \\
            UniDexGrasp++ \cite{wan2023unidexgrasp++} & Joint Values & Robot-centric & \xmark & \cmark \cmark & \xmark & \cmark & \cmark & \xmark \\
            UniGrasp \cite{shao2020unigrasp} & Contact Point & Object-centric & \cmark & \xmark & \cmark & \xmark & \xmark & \xmark \\
            GeoMatch \cite{attarian2023geometry} & Contact Point & Object-centric & \cmark & \xmark & \cmark & \cmark & \cmark & \xmark \\
            GenDexGrasp \cite{li2023gendexgrasp} & Contact Map & Object-centric & \cmark & \xmark & \cmark & \xmark & \cmark & \xmark \\
            ManiFM \cite{xu2024manifoundation} & Contact Map & Object-centric & \cmark & \xmark & \cmark & \xmark & \xmark & Contact Region \\
            \textbf{DRO-Grasp (Ours)} & $\mathcal{D(R,O)}$ & Interaction-centric & \cmark & \cmark & \cmark & \cmark & \cmark & Palm Orientation \\ 
            \bottomrule
        \end{tabular}
    }
    \caption{Dexterous grasp method comparison.}\label{tab:relatedgrasp}
    \vspace{-15pt}
\end{table*}

To overcome the limitations of both paradigms, we propose $\mathcal{D(R,O)}$, a unified representation that captures the relationship between the robotic hand’s grasp shape and the object. $\mathcal{D(R,O)}$ encapsulates both the articulated structure of the robot hand and the object’s geometry, enabling direct inference of kinematically valid and stable grasps that generalize across various shapes and robot embodiments.

Given the point clouds of both an open robotic hand and the object, our network predicts the $\mathcal{D(R,O)}$ representation, a matrix that encodes the relative distances between the point clouds of the object and the robotic hand in the desired grasping pose. Using this representation, we apply a multilateration method~\cite{norrdine2012algebraic} to estimate the robot's point cloud at the predicted pose, allowing us to compute the 6D pose of each hand link in the world frame and ultimately determine the joint configurations. To encode robotic hands, we propose a configuration-invariant pretraining method that learns the inherent alignment between various hand configurations, promoting grasp generation performance and cross-embodiment generalization. We validate the effectiveness of our approach through extensive experiments in both simulation and real-world settings. Our model achieves an average success rate of 87.53\% in simulation across three dexterous robotic hands and an 89\% success rate in real-robot experiments, demonstrating its robustness and versatility.

In conclusion, our primary contributions are as follows:
\begin{enumerate}
    \item We introduce a novel representation, $\mathcal{D(R,O)}$ for dexterous grasping tasks. This interaction-centric formulation transcends conventional robot-centric and object-centric paradigms, facilitating robust generalization across diverse robotic hands and objects.
    \item We propose a configuration-invariant pretraining approach with contrastive learning, establishing inherent alignment across varying configurations of robotic hands. This unified task can facilitate valid grasp generation and cross-embodiment feature alignment.
    \item We perform extensive experiments in both simulation environments and real-world settings, validating the efficacy of our proposed representation and framework in grasping novel objects with multiple robotic hands.
\end{enumerate}
\section{Related Work}
\subsection{Learning-based Robotic Dexterous Grasping}
Learning to grasp with dexterous robotic hands has received increasing attention in recent years. Table~\ref{tab:relatedgrasp} presents a comparison of various dexterous grasping approaches. The task of dexterous grasping is particularly challenging due to the high degree of skill required. One line of works adopts \textit{object-centric} representation for dexterous grasping, such as contact points~\cite{shao2020unigrasp,attarian2023geometry,li2021end} and contact maps~\cite{li2023gendexgrasp,xu2024manifoundation,morrison2018closing,varley2015generating}. The system infers grasp poses and joint configurations by solving inverse kinematics based on predicted contact points or maps. While object-centric representations efficiently encode object shapes and grasp information, they often face limitations in accuracy and computational efficiency when inferring precise robotic grasp configurations.

Another line of research focuses on \textit{robot-centric} approaches, which directly infer robot poses~\cite{xu2021adagrasp} or joint angles~\cite{wan2023unidexgrasp++}. These methods are typically easier to execute as they directly infer joint values for dexterous hands. However, reinforcement learning in high-dimensional action spaces often suffers from sample inefficiency and is commonly trained in simulation. The sim-to-real gap~\cite{zhao2020sim, wang2024penspin} further complicates policy transfer, with some works only reporting simulation results without real-world validation. To address these challenges, several methods leverage object-centric features~\cite{mandikal2021learning}, human grasp priors~\cite{mandikal2022dexvip}, or interaction bisector surface as the observation representation~\cite{she2022learning} to accelerate policy generation. Despite these efforts, the robot-centric approach limits the generalizability across different robotic embodiments. To combine the strengths of both approaches, we propose learning the relative distances as an interaction relationship between the robot and object for dexterous robotic grasping. This approach achieves robust real-world grasping performance and cross-embodiment generalization.

\subsection{Learning Robotic Hand Features}
To achieve cross-embodiment grasping, the grasping model needs to be aware of the descriptions of robotic hands.
UniGrasp~\cite{shao2020unigrasp} learned an embedding space with auto-encoder after transferring robot hands to point clouds. AdaGrasp~\cite{xu2021adagrasp} used a 3D TSDF volume~\cite{Newcombe2011KinectFusion} to encode the robot hand.   ManiFM~\cite{xu2024manifoundation} and GeoMatch~\cite{attarian2023geometry} encoded the robot hand by directly inputting its point cloud representation. All these approaches rely on specific robotic hand configurations. In contrast, we propose a configuration-invariant pretraining approach using contrastive learning to learn correspondences across different hand configurations, promoting effective grasp generation and cross-embodiment feature alignment.
\section{Method\label{sec:method}}
\begin{figure*}[t] \centering
    \includegraphics[width=0.98 \linewidth]{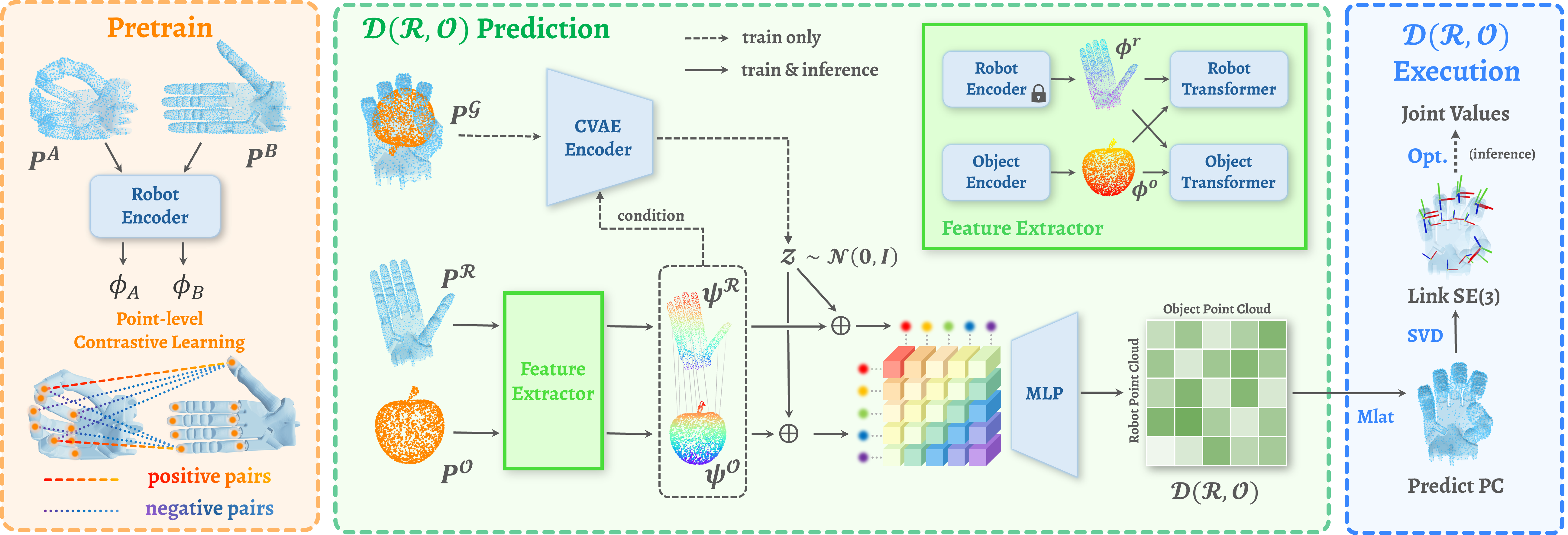}
    \caption{Overview of $\mathcal{D(R, O)}$ framework: We first pretrain the robot encoder with the proposed configuration-invariant pretraining method. Then, we predict the $\mathcal{D(R, O)}$ representation between the robot and object point cloud. Finally, we extract joint values from the $\mathcal{D(R, O)}$ representation.} \label{fig:pipeline}
    \vspace{-15pt}
\end
{figure*}

\begin{figure}[t] \centering
    \includegraphics[width=0.95 \linewidth]{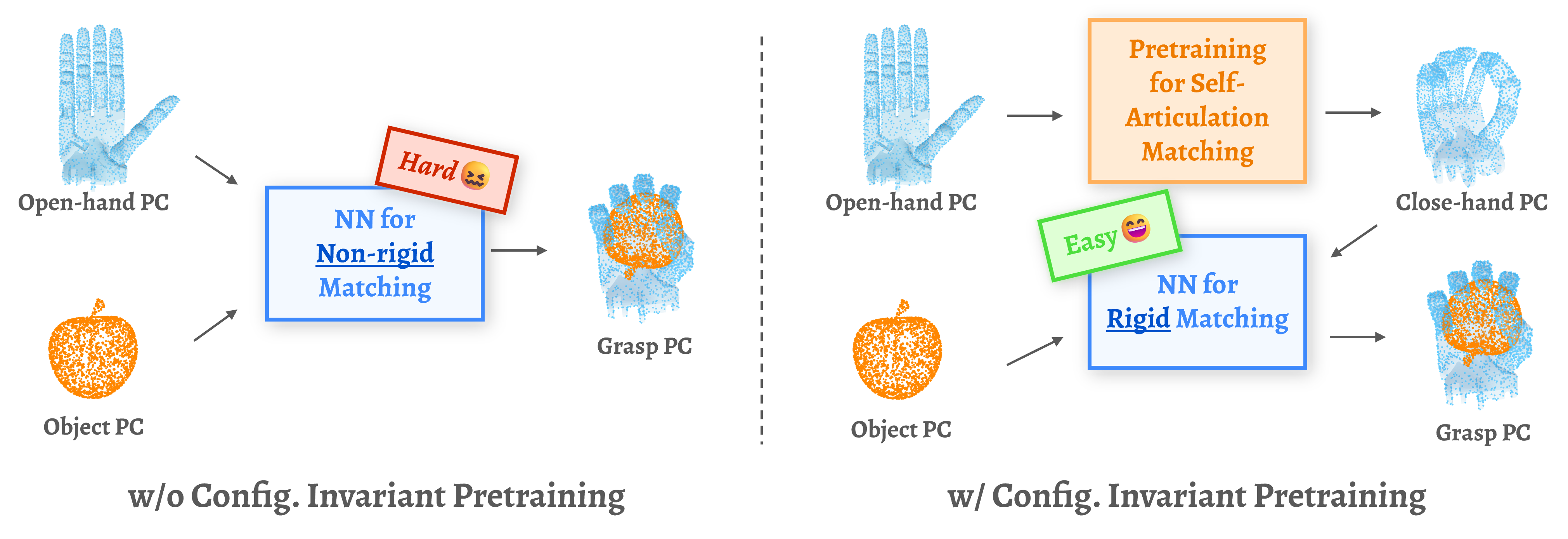}
    \vspace{-5pt}
    \caption{Motivation for configuration-invariant pretraining.} \label{fig:pretrain}
    \vspace{-15pt}
\end{figure}

Given the object point cloud and the robot hand URDF file, our goal is to generate dexterous and diverse grasping poses that generalize across various objects and robot hands. Fig.~\ref{fig:pipeline} provides an overview of our proposed method.

\textbf{Method Overview.}
First, we design an encoder network to learn representations from the point clouds of both the robot and the object. The robot encoder network is pretrained using our proposed configuration-invariant pretraining method (Sec.~\ref{subsec:pretrain}), which facilitates the learning of efficient robot embedding. Next, a CVAE model is used to predict the $\mathcal{D(R, O)}$ representation, a point-to-point distance matrix between the robotic hand at its grasp pose and the object, to implicitly present the grasp pose (Sec.~\ref{subsec:arch}). From the $\mathcal{D(R, O)}$ representation, we derive the 6D pose for each link, which serves as the optimization target for determining the joint values. This optimization process is notably straightforward and efficient (Sec.~\ref{subsec:optim}).

\subsection{Configuration-Invariant Pretraining} \label{subsec:pretrain}
Learning dexterous grasping involves understanding the spatial relationships between the robot hand and the object. The objective is to match the robot hand in a specific configuration with the object. However, this is challenging because the local geometric features of a point in the open-hand configuration may not align with those in the grasp configuration due to significant variations during articulation.

To address this, we break the problem into two simpler components: (1) self-articulation matching, which implicitly determines the joint values for the grasp configuration, and (2) wrist pose estimation. As shown in Fig.~\ref{fig:pretrain}, leveraging configuration-invariant pretraining, we train the neural network to understand the self-articulation alignment across different configurations, thereby facilitating the matching process between the robot hand and the object.

Specifically, to establish correspondence between open-hand and close-hand configurations, we randomly sample a successful grasp $q_{\mathcal{A}}$ from the dataset and compute the corresponding canonical configuration $q_{\mathcal{B}}$ with a similar wrist pose. To align point clouds with the same point order, we uniformly sample points on the surface of each link for each robotic hand, storing the resulting point clouds as $\left\{\mathbf{P}_{\ell_i}\right\}_{i=1}^{N_{\ell}}$, where $N_{\ell}$ is the number of links. We define a point cloud forward kinematics model, $\text{FK}\left(q, \left\{\mathbf{P}_{\ell_i}\right\}_{i=1}^{N_{\ell}}\right)$ to map joint configurations to point clouds. Using this model, we obtain two point clouds $\mathbf{P}^{\mathcal{A}}, \mathbf{P}^{\mathcal{B}} \in \mathbb{R}^{N_{\mathcal{R}} \times 3}$, representing these two joint configurations. Here, $N_{\mathcal{R}}$ is the number of points in the robot point cloud, set to 512 in practice. 

These point clouds are passed through the encoder network (as described in Sec.~\ref{subsec:arch}) to produce point-wise features $\boldsymbol{\phi}^{\mathcal{A}}, \boldsymbol{\phi}^{\mathcal{B}} \in \mathbb{R}^{N_{\mathcal{R}} \times D}$, where $D = 512$ is the feature dimension. The model applies point-level contrastive learning, aligning embeddings of positive pairs—points with the same index in both clouds—while separating negative pairs, weighted by the Euclidean distance in $\mathbf{P}^{\mathcal{B}}$. This process ensures that the features corresponding to the same positions on the robot hand remain consistent across different joint configurations. We define the resulting contrastive loss as:
\begin{align}
    \mathcal{L}_{p} &= -\frac{1}{N_{\ell}}\sum_{i}\log \left[ 
        \frac
            {\exp \left(
                \left<
                    \boldsymbol{\phi}_{i}^{\mathcal{A}}, \boldsymbol{\phi}_{i}^{\mathcal{B}}
                \right> / \tau
            \right)}
            {\sum_{j} \omega_{ij} \exp \left(
                \left<
                    \boldsymbol{\phi}_{i}^{\mathcal{A}}, \boldsymbol{\phi}_{j}^{\mathcal{B}}
                \right> / \tau
            \right)}
    \right],
    \\
    \omega_{ij} &= \begin{cases}
        \frac
            {\tanh \left(
                \lambda\left\|
                    p_{i}^{\mathcal{B}}-p_{j}^{\mathcal{B}}
                \right\|_{2}
            \right)}
            {\max\left(
                \tanh \left(
                    \lambda\left\|
                        p_i^{\mathcal{B}}-p_j^{\mathcal{B}}
                    \right\|_2
                \right)
            \right)}, & \text{if } i \neq j \\
        1, & \text{if } i = j
    \end{cases},
\end{align}
where $\left<\cdot, \cdot\right>$ denotes the cosine similarity between two vectors, $p_i^{\mathcal{B}}$ represents the $i$-th point position in $\mathbf{P}^{\mathcal{B}}$. For the hyperparameters, we set $\tau = 0.1$ and $\lambda = 10$ in practice.

\subsection{\texorpdfstring{$\mathcal{D(R, O)}$}{D(R, O)} Prediction} \label{subsec:arch}
Given the wrist pose which can be either randomly sampled or user-specified, we obtain an open-hand configuration $q_{\text{init}}$. The robot point cloud under $q_{\text{init}}$ is $\mathbf{P}^{\mathcal{R}} = \text{FK}\left(q_{\text{init}}, \left\{\mathbf{P}_{\ell_i}\right\}_{i=1}^{N_{\ell}}\right)\in \mathbb{R}^{N_{\mathcal{R}} \times 3}$, and the object point cloud is $\mathbf{P}^{\mathcal{O}} \in \mathbb{R}^{N_{\mathcal{O}} \times 3}$, where the number of points $N_{\mathcal{O}}$ is also 512 in practice. The objective of our neural network is to predict the point-to-point distance matrix $\mathcal{D(R, O)} \in \mathbb{R}^{N_{\mathcal{R}} \times N_{\mathcal{O}}}$, where both point clouds share the same origin. 

\textbf{Point Cloud Feature Extraction} 
We begin by extracting point cloud embeddings using two encoders, $f_{\theta_{\mathcal{R}}}(\mathbf{P}^{\mathcal{R}})$ and $f_{\theta_{\mathcal{O}}}(\mathbf{P}^{\mathcal{O}})$, which share the same architecture. Specifically, we use a modified DGCNN~\cite{wang2019dynamic} to better capture local structures and integrate global information (Appendix~\ref{app:arch_encoder}). The robot encoder is initialized with pretrained parameters, using the method described in Sec.~\ref{subsec:pretrain}, and remains frozen during training. These encoders extract point-wise features, $\boldsymbol{\phi}^{\mathcal{R}}$ and $\boldsymbol{\phi}^{\mathcal{O}}$ from the robot and object point clouds:
\begin{align}
    \boldsymbol{\phi}^{\mathcal{R}} = f_{\theta_{\mathcal{R}}}(\mathbf{P}^{\mathcal{R}}) &\in \mathbb{R}^{N_{\mathcal{R}} \times D},
    \\
    \boldsymbol{\phi}^{\mathcal{O}} = f_{\theta_{\mathcal{O}}}(\mathbf{P}^{\mathcal{O}}) &\in \mathbb{R}^{N_{\mathcal{O}} \times D}.
\end{align}

To establish correspondences between the robot and object features, we apply two multi-head cross-attention transformers~\cite{vaswani2017attention} (Appendix~\ref{app:arch_tf}), $g_{\theta_{\mathcal{R}}}(\boldsymbol{\phi}^{\mathcal{R}}, \boldsymbol{\phi}^{\mathcal{O}})$ and $g_{\theta_{\mathcal{O}}}(\boldsymbol{\phi}^{\mathcal{O}}, \boldsymbol{\phi}^{\mathcal{R}})$. These transformers integrate the relationships between the two feature sets, embedding correspondence information. This process maps the robot and object features to two sets of correlated features, $\boldsymbol{\psi}^{\mathcal{R}}$ and $\boldsymbol{\psi}^{\mathcal{O}}$:
\begin{align}
    \boldsymbol{\psi}^{\mathcal{R}} = g_{\theta_{\mathcal{R}}}(\boldsymbol{\phi}^{\mathcal{R}}, \boldsymbol{\phi}^{\mathcal{O}}) &+ \boldsymbol{\phi}^{\mathcal{R}} \in \mathbb{R}^{N_{\mathcal{R}} \times D},
    \\
    \boldsymbol{\psi}^{\mathcal{O}} = g_{\theta_{\mathcal{O}}}(\boldsymbol{\phi}^{\mathcal{O}}, \boldsymbol{\phi}^{\mathcal{R}}) &+ \boldsymbol{\phi}^{\mathcal{O}} \in \mathbb{R}^{N_{\mathcal{O}} \times D}.
\end{align}

\textbf{CVAE-based $\boldsymbol{\mathcal{D(R, O)}}$ Prediction}~\label{sec:cvae}
To achieve cross-embodiment grasp diversity, we employ a Conditional Variational Autoencoder (CVAE)~\cite{sohn2015learning} network to capture variations across numerous combinations of hand, object, and grasp configurations. The CVAE encoder $f_{\theta_{\mathcal{G}}}$ takes the robot and object point clouds under the grasp pose $\mathbf{P}^{\mathcal{G}} \in \mathbb{R}^{(N_{\mathcal{R}} + N_{\mathcal{O}}) \times 3}$, along with the learned features $(\boldsymbol{\psi}^{\mathcal{R}}, \boldsymbol{\psi}^{\mathcal{O}})$ , resulting in an input shape of $(N_{\mathcal{R}} + N_{\mathcal{O}}) \times (3 + D)$. The encoder outputs the latent variable $z\in \mathbb{R}^{d}$, set as $d = 64$ in practice. We concatenate $z$ with extracted features $\boldsymbol{\psi}^{\mathcal{R}}$ and $\boldsymbol{\psi}^{\mathcal{O}}$, converting the feature to $\widehat{\boldsymbol{\psi}}^{\mathcal{R}}_i, \widehat{\boldsymbol{\psi}}^{\mathcal{O}}_j \in \mathbb{R}^{N_{\mathcal{O}} \times (D + d)}$.

The same kernel function $\mathcal{K}$ as~\cite{eisner2024deep} is adopted, which possesses the properties of non-negativity and symmetry, to predict pair-wise distance $r_{ij} = \mathcal{K}(\widehat{\boldsymbol{\psi}}_i^{\mathcal{R}}, \widehat{\boldsymbol{\psi}}_j^{\mathcal{O}}) \in \mathbb{R}^+$ under the grasp pose:
\begin{equation}
    \small{
        \mathcal{K}(\widehat{\boldsymbol{\psi}}_i^{\mathcal{R}}, \widehat{\boldsymbol{\psi}}_j^{\mathcal{O}}) = \sigma\left(
            \frac{1}{2}\ \mathcal{N}_\theta\left(
                \widehat{\boldsymbol{\psi}}_i^{\mathcal{R}}, \widehat{\boldsymbol{\psi}}_j^{\mathcal{O}}
            \right) + \frac{1}{2}\ \mathcal{N}_\theta\left(
                \widehat{\boldsymbol{\psi}}_j^{\mathcal{O}}, \widehat{\boldsymbol{\psi}}_i^{\mathcal{R}}
            \right)
        \right)
    },
\end{equation}
where $\sigma$ denotes the $\text{softplus}$ function, and $\mathcal{N}_\theta$ is an MLP, which takes in the feature of $\mathbb{R}^{N_{\mathcal{O}} \times (2D + 2d)}$ and outputs a positive number (Appendix~\ref{app:arch_mlp}). By calculating on all $(\widehat{\boldsymbol{\psi}}_i^{\mathcal{R}}, \widehat{\boldsymbol{\psi}}_j^{\mathcal{O}})$ pairs, we obtain the complete $\mathcal{D(R,O)}$ representation:
\begin{equation}
    \mathcal{D(R,O)} = \left[\  
        \begin{matrix}
            \mathcal{K}(\widehat{\boldsymbol{\psi}}_1^{\mathcal{R}}, \widehat{\boldsymbol{\psi}}_1^{\mathcal{O}}) & \cdots & \mathcal{K}(\widehat{\boldsymbol{\psi}}_1^{\mathcal{R}}, \widehat{\boldsymbol{\psi}}_{{N_\mathcal{O}}}^{\mathcal{O}}) \\
            \vdots & \ddots & \vdots \\
            \mathcal{K}(\widehat{\boldsymbol{\psi}}_{{N_\mathcal{R}}}^{\mathcal{R}}, \widehat{\boldsymbol{\psi}}_1^{\mathcal{O}}) & \cdots & \mathcal{K}(\widehat{\boldsymbol{\psi}}_{{N_\mathcal{R}}}^{\mathcal{R}}, \widehat{\boldsymbol{\psi}}_{{N_\mathcal{O}}}^{\mathcal{O}})
        \end{matrix}
    \ \right].
\end{equation}

\begin{table*}[htbp]
    \centering
    \renewcommand\arraystretch{1.25}
    \captionsetup{justification=centering, singlelinecheck=false}
    \resizebox{\textwidth}{!}{
        \begin{threeparttable}
            \begin{tabular}{c|ccc|c|ccc|ccc}
                \toprule
                \multirow{2}{*} {\textbf{Method}} 
                & \multicolumn{4}{c|}{\textbf{Success Rate (\%) $\uparrow$}}
                & \multicolumn{3}{c|}{\textbf{Diversity (rad.) $\uparrow$}}
                & \multicolumn{3}{c}{\textbf{Efficiency (sec.) $\downarrow$}}
                \\ 
                \cline{2-11} 
                & Barrett & Allegro & ShadowHand & Avg.
                & Barrett & Allegro & ShadowHand
                & Barrett & Allegro & ShadowHand
                \\ \hline
                DFC~\cite{dfc}
                    & 86.30 & 76.21 & 58.80 & 73.77
                    & \textbf{0.532} & \textbf{0.454} & 0.435
                    & $>$1800 & $>$1800 & $>$1800
                \\
                GenDexGrasp~\cite{li2023gendexgrasp}
                    & 67.00 & 51.00 & 54.20 & 57.40
                    & 0.488 & 0.389 & 0.318
                    & 14.67 & 25.10 & 19.34
                \\
                ManiFM~\cite{xu2024manifoundation}
                    & - & 42.60 & - & 42.60
                    & - & 0.288 & -
                    & - & 9.07 & -
                \\ \hline
                DRO-Grasp (w/o pretrain)
                    & 87.20 & 82.70 & 46.70 & 72.20
                    & \textbf{0.532} & 0.448 & 0.429
                    & \textbf{0.49} & \textbf{0.47} & \textbf{0.98}
                \\ 
                \textbf{DRO-Grasp (Ours)}
                    & \textbf{87.30} & \textbf{92.30} & \textbf{83.00} & \textbf{87.53}
                    & 0.513 & 0.397 & \textbf{0.441}
                    & \textbf{0.49} & \textbf{0.47} & \textbf{0.98}
                \\ 
                \bottomrule
            \end{tabular}
        \end{threeparttable}
    }
    \caption{Overall comparison with baselines.}
    \label{tab:result}
    \vspace{-15pt}
\end{table*}

\subsection{Grasp Configuration Generation from \texorpdfstring{$\mathcal{D(R, O)}$}{D(R, O)}}\label{sec:graspconfig}
Given the predicted $\mathcal{D(R,O)}$, we discuss how to generate the grasp joint values to grasp the object. We first calculate the robot grasp point cloud, then estimate each link's 6D pose based on the joint clouds. The system calculates the joint values by matching each link's 6D pose.

\textbf{Robotic Grasp Pose Point Cloud Generation} For a given point $p_i^{\mathcal{R}}$, the $i$-th row of $\mathcal{D(R,O)}$ denotes the distances from this robot grasp point to all points in the object point cloud. Given the object point cloud, the multilateration method~\cite{norrdine2012algebraic} positions the robot point cloud. This positioning technique determines the location of a point $p_i'^{\mathcal{R}}$ by solving the least-squares optimization problem based on distances from multiple reference points:
\begin{equation}
    p_i'^{\mathcal{R}} = \underset{p_i^{\mathcal{R}}}{\arg\min} \sum_{j=1}^{N_\mathcal{O}} \left(
        \| p_i^{\mathcal{R}} - p_j^{\mathcal{O}} \|_2^2 - \mathcal{D(R,O)}_{ij}^2 
    \right)^2.
\end{equation}

As shown in~\cite{zhou2009efficient}, this problem has a closed-form solution, and by using the implementation from~\cite{eisner2024deep}, we can directly compute $p_i'^{\mathcal{R}}$. Repeating this process for each row of $\mathcal{D(R,O)}$ yields the complete predicted robot point cloud $\mathbf{P}^{\mathcal{P}}$ in the grasp pose. In 3D space, we can determine a point's position by measuring its relative distances to just 4 other points. Our $\mathcal{D(R,O)}$ representation provides $N_{\mathcal{O}} (=512)$ relative distances, enhancing robustness to prediction errors.

\textbf{6D Pose Estimation of Links}
Directly solving inverse kinematics and getting the joint values from a point cloud is not a trivial task. We first compute the 6D pose of each link in the world frame. As described in Sec.~\ref{subsec:pretrain}, we store the point cloud for each link, $\left\{\mathbf{P}_{\ell_i}\right\}_{i=1}^{N_{\ell}}$. Given the predicted grasp point cloud $\left\{\mathbf{P}_{\ell_i}^{\mathcal{P}}\right\}_{i=1}^{N_{\ell}}$, we calculate the 6D pose of each link using rigid body registration techniques:
\begin{equation}
\boldsymbol{\mathcal{T}^*}=(\mathbf{x}_i^*, \mathbf{R}^*_i) =\argmin_{ (\mathbf{x}_i, \mathbf{R}_i) } \|\mathbf{P}_{\ell_i}^{\mathcal{P}}-\mathbf{P}_{\ell_i} (\mathbf{x}_i, \mathbf{R}_i) \|^2,
\label{eqn:poseE}
\end{equation}
where $\mathbf{x}_i$ and $\mathbf{R}_i$ represent the translation and rotation of the $i$-th link, respectively. This computation can be directly performed using singular value decomposition (SVD).

\textbf{Joint Configuration Optimization} \label{subsec:optim}
After predicting the 6D pose for each link, our objective is to optimize the joint values to align the translation of each link with the predicted result. During the inference phase, we initialize with $q_{init}$ and iteratively refine the solution through the following optimization problem using CVXPY~\cite{diamond2016cvxpy}:
\begin{align}
    \min_{\delta \boldsymbol{q}} &\left(
        \sum_{i=1}^{N_\ell} \left\| 
            \mathbf{x}_i + \frac
                {\partial\mathbf{x}_i(\boldsymbol{q})}
                {\partial{\boldsymbol{q}}}
            \delta \boldsymbol{q} - \mathbf{x}_i^*
        \right\|_{2} 
    \right),
    \\
    \text{s.t. }&
    \boldsymbol{q} + \delta \boldsymbol{q} \in [\boldsymbol{q}_{min},\boldsymbol{q}_{max}],
    \ |\delta \boldsymbol{q}| \leq \varepsilon_q.
\end{align}

In each iteration, the system computes the delta joint values $\delta \boldsymbol{q}$ by minimizing the objective function and updates the joint values as $\boldsymbol{q} \leftarrow \boldsymbol{q} + \delta \boldsymbol{q}$. Here, $\mathbf{x}_i$ represents the current link translation, $[\boldsymbol{q}_{min},\boldsymbol{q}_{max}]$ denotes the joint limits, and $\varepsilon_q=0.5$ is the maximum allowable step size. The optimization process can be efficiently parallelized, stably achieving convergence within one second, even for a 6+22 DoF ShadowHand.

\subsection{Loss Function}
Notably, the computation from $\mathcal{D(R,O)}$ representation to the 6D pose $\boldsymbol{\mathcal{T}}^*$ shown in Eqn.~\ref{eqn:poseE} is entirely matrix-based, ensuring differentiability for loss backpropagation and computational efficiency.

The training objectives of the whole network include four parts, including the prediction of $\mathcal{D(R,O)}$ and $\boldsymbol{\mathcal{T}}$, the suppression of penetration, and the KL divergence of the CVAE latent variable (described in~\ref{sec:cvae}):
\begin{equation}
    \begin{aligned}
        \mathcal{L} &= \lambda_{\mathcal{D}}\mathcal{L_{\text{L1}}}\left(
            \mathcal{D(R,O)}, \mathcal{D(R,O)}^{\text{GT}}
        \right)
        \\
        &+ \lambda_{\mathcal{T}}\frac{1}{N_\ell}\sum_{i=1}^{N_\ell}\mathcal{L}_{\ell_i}
        + \lambda_{\mathcal{P}}\left|
            \mathcal{L_\text{P}}(\mathbf{P}^{\mathcal{T}}, \mathbf{P}^{\mathcal{O}})
        \right|
        \\
        &+ \lambda_{KL}\mathcal{D}_{KL}\left(
            f_{\theta_{\mathcal{G}}}(\mathbf{P}^{\mathcal{G}}, \boldsymbol{\psi}^{\mathcal{R}}, \boldsymbol{\psi}^{\mathcal{O}})\ \|\ \mathcal{N}(0, I)
        \right),
    \end{aligned}
\end{equation}
where $\lambda_{\mathcal{D}}$, $\lambda_{\mathcal{T}}$, $\lambda_{\mathcal{P}}$, $\lambda_{KL}$ are hyperparameters for loss weights. The superscript ``GT'' refers to the ground truth annotations. $\mathcal{N}(0, I)$ is a standard Guassian distribution, and $\mathbf{P}^{\mathcal{T}}$ is the robot point cloud under the $\boldsymbol{\mathcal{T}^*}$ described in ~\ref{sec:graspconfig}. $\mathcal{L_\text{P}}$ computes the sum of the negative values of the signed distance function (SDF) of $\mathbf{P}^{\mathcal{T}}$ to $\mathbf{P}^{\mathcal{O}}$ to penalize any penetration between the robot hand and the object, and $\mathcal{L_\ell}$ computes the difference between two 6D poses:
\begin{equation}
    \mathcal{L}_{\ell_i} = \| \mathbf{x}_i^* - \mathbf{x}_i^{\text{GT}} \|_2 + \arccos\left(
        \frac{\mathrm{tr}(\mathbf{R}_i^\mathrm{*T}\mathbf{R}_i^{\text{GT}}) - 1}{2}
    \right).
\end{equation}

\section{Experiments}
\begin{figure}[t] \centering
    \includegraphics[width=0.98 \linewidth]{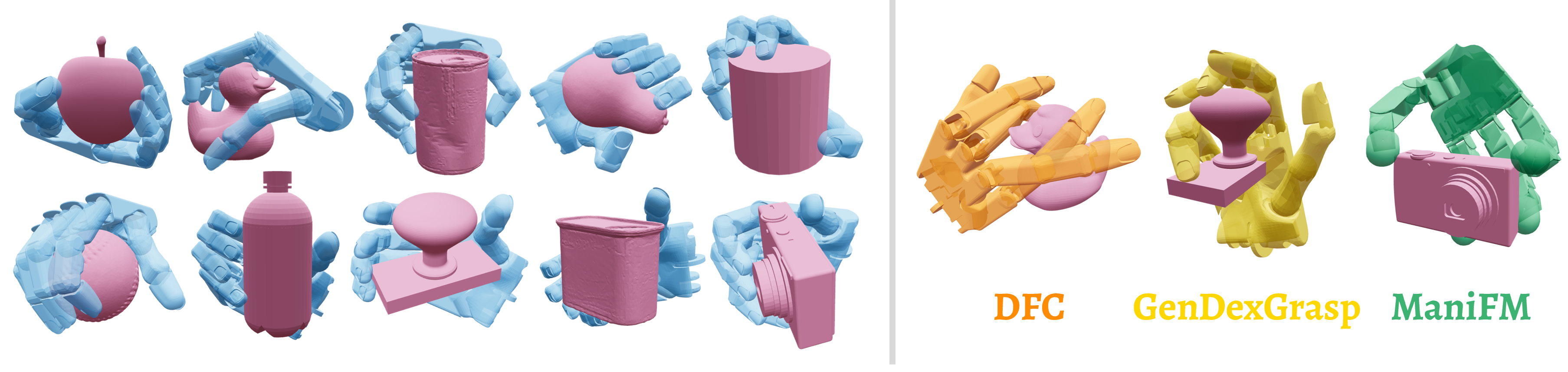}
    \caption{Visualization of generated grasps, compared to typical failure cases from existing approaches.}
    \vspace{-14pt}
    \label{fig:grasp}
\end{figure}

In this section, we perform a series of experiments aimed at addressing the following questions (Q1-Q7):

\begin{itemize}[leftmargin=5pt]
    \item[] Q1: How successful are our generated grasps?
    \item[] Q2: Does our unified model train on multi-embodiment outperform models trained on single embodiments? 
    \item[] Q3: How diverse are our generated grasps?
    \item[] Q4: How well does our pretraining learn configuration-invariant representations, and can this be transferred across different embodiments?
    \item[] Q5: How robust is our approach with partial object point cloud input?
    \item[] Q6: How does our method perform in real-world settings?  
    \item[] Q7: Can our method generalize to novel robot hands? (Appendix~\ref{app:zeroshot})
\end{itemize}

\subsection{Evaluation Metric\label{sec:evalmetric}}
\textbf{Success Rate:} We evaluate the success of grasping by determining whether the force closure condition is satisfied. To implement this evaluation criterion, we used the Isaac Gym simulator~\cite{liang2018gpu}. A simple grasp controller is applied to execute the predicted grasps in the simulation (Appendix~\ref{app:controller}). Following the metric in~\cite{li2023gendexgrasp}, we sequentially apply forces along six orthogonal directions, each for a duration of 1 second. The grasp is considered successful if the object's resultant displacement remains below 2 cm after all six directional forces have been applied.

\textbf{Diversity:} Grasp diversity is quantified by calculating the standard deviation of the joint values (including 6 floating wrist DoF) across all successful grasps. 

\textbf{Efficiency:} The computational time required to achieve a grasp is measured, encompassing both network inference and the subsequent optimization steps.

\subsection{Dataset} \label{subsec:dataset}
\vspace{-1mm}
We utilized a subset of the CMapDataset~\cite{li2023gendexgrasp} (See Appendix~\ref{app:filter} for the filtering process). After filtering, 24,764 valid grasps remained. We adopt three robots from the dataset: Barrett (3-finger), Allegro (4-finger), and ShadowHand (5-finger). Each grasp defines its associated object, robot, and grasp configurations. We retain the same training and test dataset splits as in the CMapDataset dataset.

\subsection{Overall Performance}
\textbf{Baselines} To answer Q1, we present a detailed comparison of $\mathcal{D(R,O)}$ against DFC~\cite{dfc}, GenDexGrasp~\cite{li2023gendexgrasp}, and ManiFM~\cite{xu2024manifoundation}, as shown in Tab.~\ref{tab:result}. This comparison includes diverse methods to address the challenge of cross-embodiment grasping from various perspectives. They were evaluated on 10 previously unseen test objects using the Barrett, Allegro, and ShadowHand robotic hands. DFC is an optimization-based approach that searches for feasible grasp configurations through iterative optimization. GenDexGrasp predicts contact heatmaps and uses optimization to determine grasp poses. ManiFM supports cross-embodiment grasping but employs a point-contact approach, which was not suitable for training on our dataset that emphasizes surface-contact methods. As a result, we can only evaluate its pretrained model of Allegro Hand for ManiFM.

Our experiments demonstrate that $\mathcal{D(R,O)}$ significantly outperformed all baselines regarding success rate across the robots by a large margin, highlighting the effectiveness of our approach. For successful grasps of our method, the average displacement remains under 2 mm, with an average rotation
below $1^\circ$, highlighting the firmness of our generated grasps. Fig.~\ref{fig:grasp} visualizes grasps generated by our method alongside typical failure grasp poses from baselines. DFC often results in unnatural poses. GenDexGrasp struggles with objects of complex shapes, frequently encountering significant penetration issues. Although ManiFM produces visually appealing grasps, its point-contact method lacks stability, lowering its success rate in simulation.

From the first two rows of Tab.~\ref{tab:cross}, we can see a slight improvement in success rates when training across multiple robots compared to training on a single hand, demonstrating the cross-embodiment generalizability of our method (Q2).

Our method significantly improves grasp generation speed. While DFC is slow in producing results and learning-based methods like GenDexGrasp and ManiFM take tens of seconds per grasp due to their complex optimization processes, our approach can generate a grasp within 1 second. This fast computation is crucial for dexterous manipulation tasks.

\begin{table}[tbp]
    \centering
    \renewcommand\arraystretch{1.25}  
    \captionsetup{justification=centering, singlelinecheck=false}
    \resizebox{\columnwidth}{!}{
        \begin{tabular}{c|ccc|ccc}
            \toprule
            \multirow{2}{*} {\textbf{Method}} 
            & \multicolumn{3}{c|}{\textbf{Success Rate (\%) $\uparrow$}} 
            & \multicolumn{3}{c}{\textbf{Diversity (rad) $\uparrow$}}
            \\
            \cline{2-7} 
            & Barrett & Allegro & ShadowHand 
            & Barrett & Allegro & ShadowHand
            \\ 
            \hline
            Single
                & 84.80 & 88.70 & 75.80
                & 0.505 & \textbf{0.435} & 0.425
            \\
            Multi
                & \textbf{87.30} & \textbf{92.30} & \textbf{83.00}
                & \textbf{0.513} & 0.397 & \textbf{0.441}
            \\ 
            \hline
            Partial
                & 84.70 & 87.60 & 81.80
                & 0.511 & 0.401 & 0.412
            \\
            \bottomrule
        \end{tabular}
    }
    \caption{Comparison under different conditions. “Single” trains on one hand, “Multi” trains on all hands, and “Partial” trains and tests on partial point clouds.}
    \label{tab:cross}
    \vspace{-8pt}
\end{table}

\subsection{Diverse Grasp Synthesis}
\begin{figure}[t] \centering
    \includegraphics[width=0.95 \linewidth]{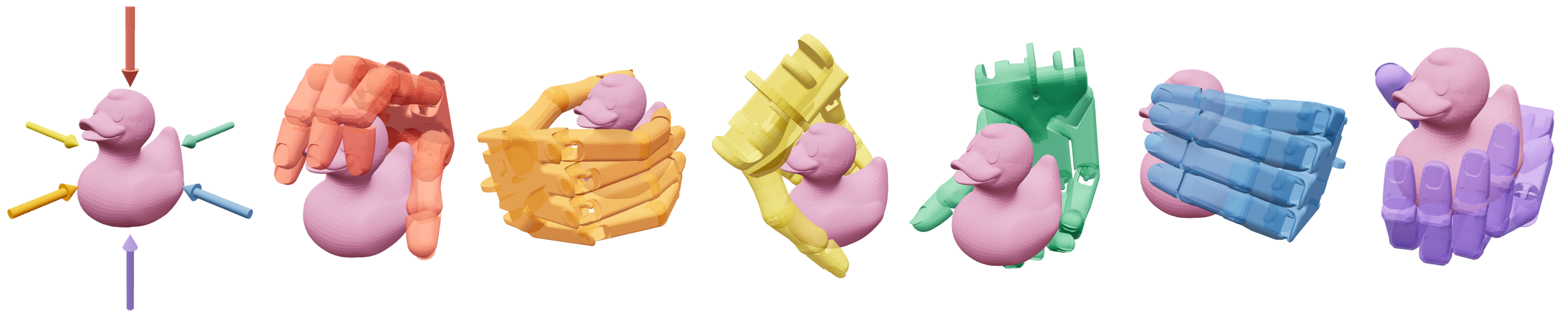}
    \caption{Diverse and pose-controllable grasp generation. The arrow refers to the input palm orientation. Arrows and hands of the same color represent corresponding input-output pairs.} \label{fig:direction}
    \vspace{-14pt}
\end{figure}

Grasping diversity includes two key aspects: the wrist pose and the finger joint values. Since the input and grasp rotations in the training data are correspondingly aligned, the model learns to implicitly map these rotations. This alignment enables the model, during inference, to generate appropriate grasps based on the specified input orientation. Fig.~\ref{fig:direction} illustrates the grasp results for six different input directions, showing that our model consistently produces feasible grasps, demonstrating the controllability of our method. Additionally, by sampling the latent variable $z \in \mathbb{R}^{64}$ from $\mathcal{N}(0, I)$, our model can generate multiple grasps in the same direction, addressing Q3. As shown in Tab.~\ref{tab:result}, the diversity of our method is highly competitive.

\begin{figure}[t] \centering
    \includegraphics[width=0.98 \linewidth]{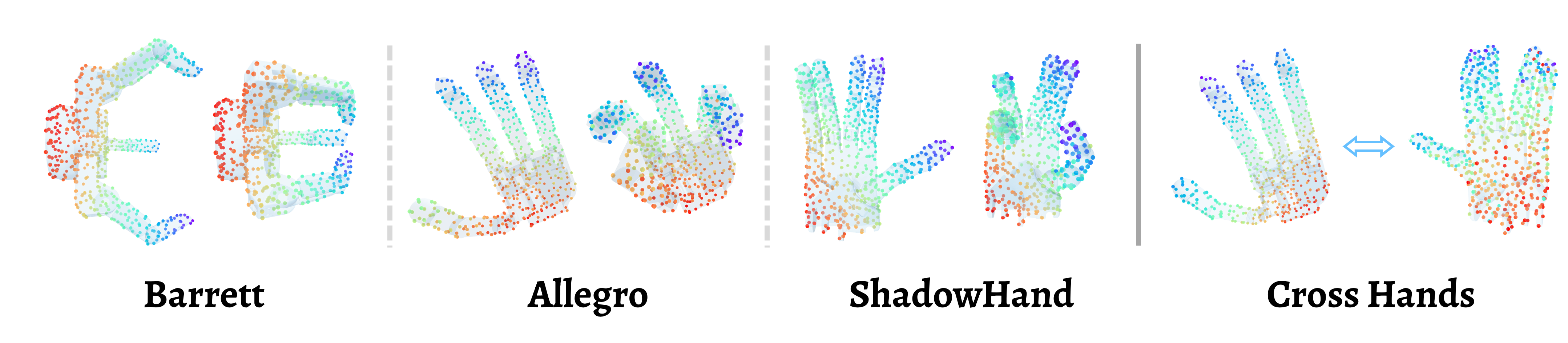}
    \vspace{0pt}
    \caption{Visualization of the pretrained point matching.} \label{fig:match}
    \vspace{-15pt}
\end{figure}

\subsection{Configuration Correspondence Learning}
As described in Sec.~\ref{subsec:pretrain}, our proposed configuration-invariant pretraining method learns an inherent alignment across varying robotic hand configurations. To answer the first part of Q4, we visualize the learned correspondence in Fig.~\ref{fig:match}, where each point in the closed-hand pose is colored according to the highest cosine similarity with its counterpart in the open-hand pose. The excellent color matching within the same hand demonstrates that the pretrained encoder successfully captures this alignment. Furthermore, strong matching across different hands highlights the transferability of features (Q4). As shown in Tab.~\ref{tab:result}, removing the pretraining parameters and training the robot encoder directly results in performance degradation across robotic hands, confirming the effectiveness of the pretrained model.

\subsection{Grasping with Partial Object Point Cloud Input}
A common challenge in real-world experiments is the noise and incompleteness of point clouds from depth cameras. Object-centric methods that rely on full object visibility often suffer performance degradation under such conditions. In contrast, the relative distance feature of $\mathcal{D(R,O)}$ allows our method to infer the robot point cloud even from partial observation without relying on complete object visibility. We validated this approach by conducting experiments, removing 50\% of the object point cloud in a contiguous region during both training and evaluation. This setup simulates the incomplete data commonly encountered in practice. As shown in the third row of Tab.~\ref{tab:cross}, even with partial point clouds, our model can successfully predict feasible grasps (Q5), indicating robustness when faced with incomplete input.

\subsection{Real-Robot Experiments}


\begin{wrapfigure}{r}{0.2\textwidth} 
    \centering
    \vspace{-12pt} 
    \includegraphics[width=\linewidth]{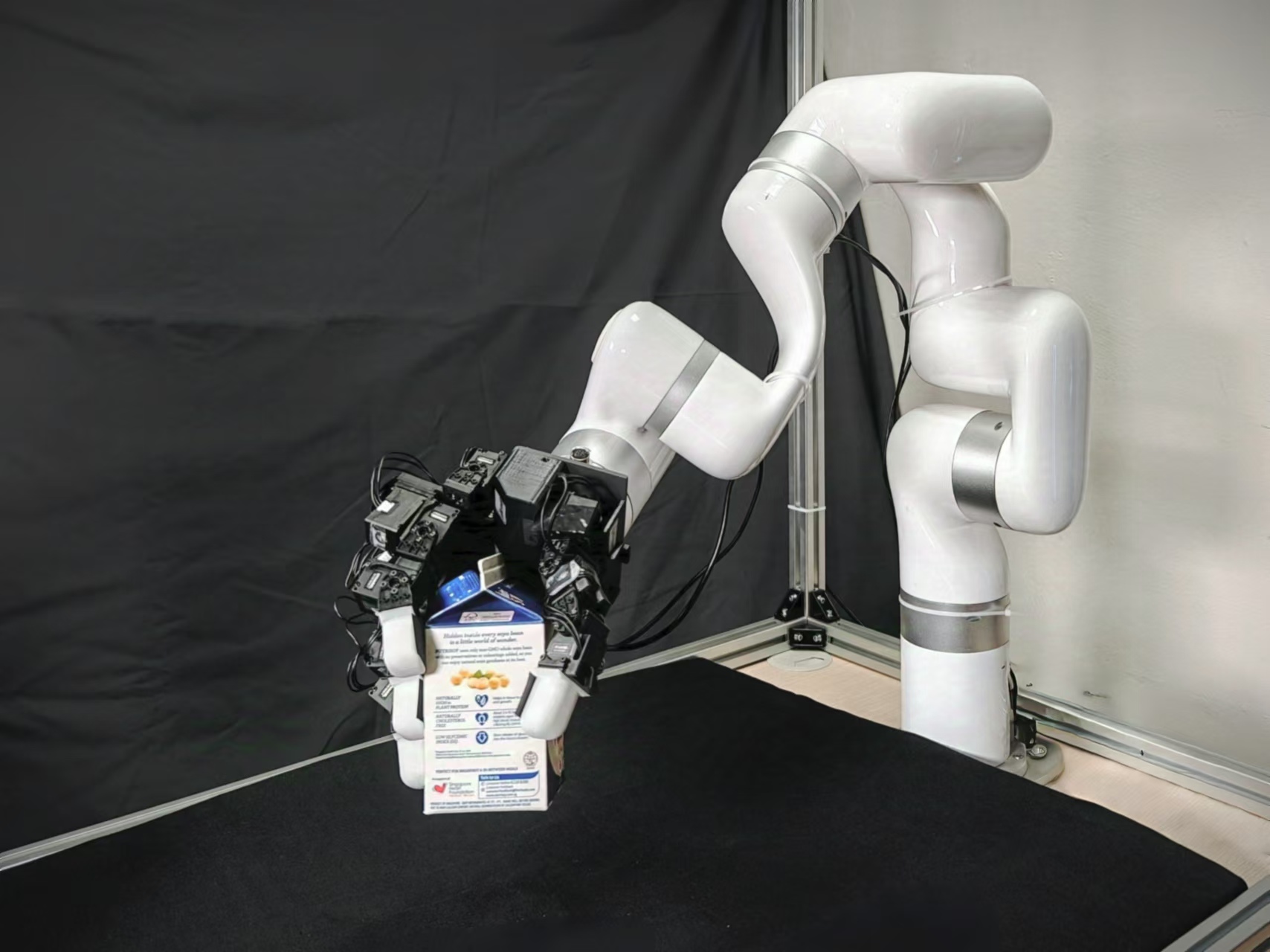}
    \vspace{-12pt} 
    \caption{Real-world experiment setting.}
    \vspace{-5pt}
    \label{fig:grasp_rw}
\end{wrapfigure}

We conducted real-robot experiments using a uFactory xArm6 robot, equipped with the LEAP Hand~\cite{shaw2023leaphand} and an overhead Realsense D435 camera, as illustrated in Fig.~\ref{fig:grasp_rw}. Our method achieved an average success rate of \textbf{89\%} across 10 novel objects, showcasing its effectiveness in dexterous grasping and generalization to previously unseen objects (Q6). Appendix~\ref{app:realworld} provides a detailed description of the experimental pipeline and quantitative results. For experiment videos, please visit our website~\href{https://nus-lins-lab.github.io/drograspweb/}{https://nus-lins-lab.github.io/drograspweb/}.

\section{Conclusion}
This work presents a new method for improving dexterous grasping by introducing the $\mathcal{D(R,O)}$ representation, which captures the essential interaction between robotic hands and objects. Unlike existing methods that rely heavily on either object or robot-specific representations, our approach bridges the gap by using a unified framework that generalizes well across different robots and object geometries. Additionally, our pretraining approach enhances the model's capacity to adapt to different hand configurations, making it suitable for a wide range of robotic systems. Experimental results confirm that our method delivers notable improvements in success rates, diversity, and computational efficiency.


{\small
\bibliographystyle{IEEEtran}
\bibliography{references}

\begin{thebibliography}{10}
\providecommand{\url}[1]{#1}
\csname url@samestyle\endcsname
\providecommand{\newblock}{\relax}
\providecommand{\bibinfo}[2]{#2}
\providecommand{\BIBentrySTDinterwordspacing}{\spaceskip=0pt\relax}
\providecommand{\BIBentryALTinterwordstretchfactor}{4}
\providecommand{\BIBentryALTinterwordspacing}{\spaceskip=\fontdimen2\font plus
\BIBentryALTinterwordstretchfactor\fontdimen3\font minus \fontdimen4\font\relax}
\providecommand{\BIBforeignlanguage}[2]{{%
\expandafter\ifx\csname l@#1\endcsname\relax
\typeout{** WARNING: IEEEtran.bst: No hyphenation pattern has been}%
\typeout{** loaded for the language `#1'. Using the pattern for}%
\typeout{** the default language instead.}%
\else
\language=\csname l@#1\endcsname
\fi
#2}}
\providecommand{\BIBdecl}{\relax}
\BIBdecl

\bibitem{roa2015grasp}
M.~A. Roa and R.~Su{\'a}rez, ``Grasp quality measures: review and performance,'' \emph{Autonomous robots}, vol.~38, pp. 65--88, 2015.

\bibitem{dfc}
T.~Liu, Z.~Liu, Z.~Jiao, Y.~Zhu, and S.-C. Zhu, ``Synthesizing diverse and physically stable grasps with arbitrary hand structures using differentiable force closure estimator,'' \emph{IEEE Robotics and Automation Letters}, vol.~7, no.~1, pp. 470--477, 2021.

\bibitem{chen2024springgrasp}
S.~Chen, J.~Bohg, and C.~K. Liu, ``Springgrasp: An optimization pipeline for robust and compliant dexterous pre-grasp synthesis,'' \emph{arXiv preprint arXiv:2404.13532}, 2024.

\bibitem{patel2024getzero}
\BIBentryALTinterwordspacing
A.~Patel and S.~Song, ``{GET-Zero}: Graph embodiment transformer for zero-shot embodiment generalization,'' 2024. [Online]. Available: \url{https://arxiv.org/abs/2407.15002}
\BIBentrySTDinterwordspacing

\bibitem{haldar2023teach}
S.~Haldar, J.~Pari, A.~Rai, and L.~Pinto, ``Teach a robot to fish: Versatile imitation from one minute of demonstrations,'' \emph{arXiv preprint arXiv:2303.01497}, 2023.

\bibitem{xu2023unidexgrasp}
Y.~Xu, W.~Wan, J.~Zhang, H.~Liu, Z.~Shan, H.~Shen, R.~Wang, H.~Geng, Y.~Weng, J.~Chen \emph{et~al.}, ``Unidexgrasp: Universal robotic dexterous grasping via learning diverse proposal generation and goal-conditioned policy,'' in \emph{Proceedings of the IEEE/CVF Conference on Computer Vision and Pattern Recognition}, 2023, pp. 4737--4746.

\bibitem{xu2023fast}
W.~Xu, W.~Guo, X.~Shi, X.~Sheng, and X.~Zhu, ``Fast force-closure grasp synthesis with learning-based sampling,'' \emph{IEEE Robotics and Automation Letters}, vol.~8, no.~7, pp. 4275--4282, 2023.

\bibitem{wan2023unidexgrasp++}
W.~Wan, H.~Geng, Y.~Liu, Z.~Shan, Y.~Yang, L.~Yi, and H.~Wang, ``Unidexgrasp++: Improving dexterous grasping policy learning via geometry-aware curriculum and iterative generalist-specialist learning,'' in \emph{Proceedings of the IEEE/CVF International Conference on Computer Vision}, 2023, pp. 3891--3902.

\bibitem{shao2020unigrasp}
L.~Shao, F.~Ferreira, M.~Jorda, V.~Nambiar, J.~Luo, E.~Solowjow, J.~A. Ojea, O.~Khatib, and J.~Bohg, ``Unigrasp: Learning a unified model to grasp with multifingered robotic hands,'' \emph{IEEE Robotics and Automation Letters}, vol.~5, no.~2, pp. 2286--2293, 2020.

\bibitem{attarian2023geometry}
M.~Attarian, M.~A. Asif, J.~Liu, R.~Hari, A.~Garg, I.~Gilitschenski, and J.~Tompson, ``Geometry matching for multi-embodiment grasping,'' in \emph{Conference on Robot Learning}.\hskip 1em plus 0.5em minus 0.4em\relax PMLR, 2023, pp. 1242--1256.

\bibitem{li2021end}
S.~Li, Z.~Li, K.~Han, X.~Li, Y.~Xiong, and Z.~Xie, ``An end-to-end spatial grasp prediction model for humanoid multi-fingered hand using deep network,'' in \emph{2021 6th International Conference on Control, Robotics and Cybernetics (CRC)}.\hskip 1em plus 0.5em minus 0.4em\relax IEEE, 2021, pp. 130--136.

\bibitem{li2023gendexgrasp}
P.~Li, T.~Liu, Y.~Li, Y.~Geng, Y.~Zhu, Y.~Yang, and S.~Huang, ``Gendexgrasp: Generalizable dexterous grasping,'' in \emph{2023 IEEE International Conference on Robotics and Automation (ICRA)}.\hskip 1em plus 0.5em minus 0.4em\relax IEEE, 2023, pp. 8068--8074.

\bibitem{xu2024manifoundation}
Z.~Xu, C.~Gao, Z.~Liu, G.~Yang, C.~Tie, H.~Zheng, H.~Zhou, W.~Peng, D.~Wang, T.~Chen, Z.~Yu, and L.~Shao, ``Manifoundation model for general-purpose robotic manipulation of contact synthesis with arbitrary objects and robots,'' 2024.

\bibitem{morrison2018closing}
D.~Morrison, P.~Corke, and J.~Leitner, ``Closing the loop for robotic grasping: A real-time, generative grasp synthesis approach,'' \emph{arXiv preprint arXiv:1804.05172}, 2018.

\bibitem{varley2015generating}
J.~Varley, J.~Weisz, J.~Weiss, and P.~Allen, ``Generating multi-fingered robotic grasps via deep learning,'' in \emph{2015 IEEE/RSJ international conference on intelligent robots and systems (IROS)}.\hskip 1em plus 0.5em minus 0.4em\relax IEEE, 2015, pp. 4415--4420.

\bibitem{wu2022learning}
A.~Wu, M.~Guo, and C.~K. Liu, ``Learning diverse and physically feasible dexterous grasps with generative model and bilevel optimization,'' \emph{arXiv preprint arXiv:2207.00195}, 2022.

\bibitem{norrdine2012algebraic}
A.~Norrdine, ``An algebraic solution to the multilateration problem,'' in \emph{Proceedings of the 15th international conference on indoor positioning and indoor navigation, Sydney, Australia}, vol. 1315, 2012.

\bibitem{xu2021adagrasp}
Z.~Xu, B.~Qi, S.~Agrawal, and S.~Song, ``Adagrasp: Learning an adaptive gripper-aware grasping policy,'' in \emph{Proceedings of the IEEE International Conference on Robotics and Automation}, 2021.

\bibitem{zhao2020sim}
W.~Zhao, J.~P. Queralta, and T.~Westerlund, ``Sim-to-real transfer in deep reinforcement learning for robotics: a survey,'' in \emph{2020 IEEE symposium series on computational intelligence (SSCI)}.\hskip 1em plus 0.5em minus 0.4em\relax IEEE, 2020, pp. 737--744.

\bibitem{wang2024penspin}
J.~Wang, Y.~Yuan, H.~Che, H.~Qi, Y.~Ma, J.~Malik, and X.~Wang, ``Lessons from learning to spin “pens”,'' \emph{arXiv:2405.07391}, 2024.

\bibitem{mandikal2021learning}
P.~Mandikal and K.~Grauman, ``Learning dexterous grasping with object-centric visual affordances,'' in \emph{2021 IEEE international conference on robotics and automation (ICRA)}.\hskip 1em plus 0.5em minus 0.4em\relax IEEE, 2021, pp. 6169--6176.

\bibitem{mandikal2022dexvip}
------, ``Dexvip: Learning dexterous grasping with human hand pose priors from video,'' in \emph{Conference on Robot Learning}.\hskip 1em plus 0.5em minus 0.4em\relax PMLR, 2022, pp. 651--661.

\bibitem{she2022learning}
Q.~She, R.~Hu, J.~Xu, M.~Liu, K.~Xu, and H.~Huang, ``Learning high-dof reaching-and-grasping via dynamic representation of gripper-object interaction,'' \emph{arXiv preprint arXiv:2204.13998}, 2022.

\bibitem{Newcombe2011KinectFusion}
R.~A. Newcombe, S.~Izadi, O.~Hilliges, D.~Molyneaux, D.~Kim, A.~J. Davison, P.~Kohi, J.~Shotton, S.~Hodges, and A.~Fitzgibbon, ``Kinectfusion: Real-time dense surface mapping and tracking,'' in \emph{2011 10th IEEE International Symposium on Mixed and Augmented Reality}, 2011, pp. 127--136.

\bibitem{wang2019dynamic}
Y.~Wang, Y.~Sun, Z.~Liu, S.~E. Sarma, M.~M. Bronstein, and J.~M. Solomon, ``Dynamic graph cnn for learning on point clouds,'' \emph{ACM Transactions on Graphics (tog)}, vol.~38, no.~5, pp. 1--12, 2019.

\bibitem{vaswani2017attention}
A.~Vaswani, ``Attention is all you need,'' \emph{Advances in Neural Information Processing Systems}, 2017.

\bibitem{sohn2015learning}
K.~Sohn, H.~Lee, and X.~Yan, ``Learning structured output representation using deep conditional generative models,'' \emph{Advances in neural information processing systems}, vol.~28, 2015.

\bibitem{eisner2024deep}
B.~Eisner, Y.~Yang, T.~Davchev, M.~Vecerik, J.~Scholz, and D.~Held, ``Deep se (3)-equivariant geometric reasoning for precise placement tasks,'' \emph{arXiv preprint arXiv:2404.13478}, 2024.

\bibitem{zhou2009efficient}
Y.~Zhou, ``An efficient least-squares trilateration algorithm for mobile robot localization,'' in \emph{2009 IEEE/RSJ International Conference on Intelligent Robots and Systems}.\hskip 1em plus 0.5em minus 0.4em\relax IEEE, 2009, pp. 3474--3479.

\bibitem{diamond2016cvxpy}
S.~Diamond and S.~Boyd, ``{CVXPY}: {A} {P}ython-embedded modeling language for convex optimization,'' \emph{Journal of Machine Learning Research}, vol.~17, no.~83, pp. 1--5, 2016.

\bibitem{liang2018gpu}
J.~Liang, V.~Makoviychuk, A.~Handa, N.~Chentanez, M.~Macklin, and D.~Fox, ``Gpu-accelerated robotic simulation for distributed reinforcement learning,'' in \emph{Conference on Robot Learning}.\hskip 1em plus 0.5em minus 0.4em\relax PMLR, 2018, pp. 270--282.

\bibitem{shaw2023leaphand}
K.~Shaw, A.~Agarwal, and D.~Pathak, ``Leap hand: Low-cost, efficient, and anthropomorphic hand for robot learning,'' \emph{Robotics: Science and Systems (RSS)}, 2023.

\bibitem{calli2017yale}
B.~Calli, A.~Singh, J.~Bruce, A.~Walsman, K.~Konolige, S.~Srinivasa, P.~Abbeel, and A.~M. Dollar, ``Yale-cmu-berkeley dataset for robotic manipulation research,'' \emph{The International Journal of Robotics Research}, vol.~36, no.~3, pp. 261--268, 2017.

\bibitem{brahmbhatt2019contactdb}
S.~Brahmbhatt, C.~Ham, C.~C. Kemp, and J.~Hays, ``Contactdb: Analyzing and predicting grasp contact via thermal imaging,'' in \emph{Proceedings of the IEEE/CVF conference on computer vision and pattern recognition}, 2019, pp. 8709--8719.

\bibitem{wang2022dexgraspnet}
R.~Wang, J.~Zhang, J.~Chen, Y.~Xu, P.~Li, T.~Liu, and H.~Wang, ``Dexgraspnet: A large-scale robotic dexterous grasp dataset for general objects based on simulation,'' \emph{arXiv preprint arXiv:2210.02697}, 2022.

\bibitem{arcode2022}
\BIBentryALTinterwordspacing
{AR Code}, ``Ar code,'' 2022, accessed: 2024-09-28. [Online]. Available: \url{https://ar-code.com/}
\BIBentrySTDinterwordspacing

\bibitem{opencv_library}
\BIBentryALTinterwordspacing
{OpenCV Team}, ``Opencv: Open source computer vision library,'' 2023, version 4.8.0, \url{https://docs.opencv.org/4.x/d9/d6c/tutorial_table_of_content_charuco.html}. [Online]. Available: \url{https://opencv.org/}
\BIBentrySTDinterwordspacing

\bibitem{chen2023easyhec}
L.~Chen, Y.~Qin, X.~Zhou, and H.~Su, ``Easyhec: Accurate and automatic hand-eye calibration via differentiable rendering and space exploration,'' \emph{IEEE Robotics and Automation Letters}, 2023.

\bibitem{wen2024foundationpose}
B.~Wen, W.~Yang, J.~Kautz, and S.~Birchfield, ``Foundationpose: Unified 6d pose estimation and tracking of novel objects,'' in \emph{Proceedings of the IEEE/CVF Conference on Computer Vision and Pattern Recognition}, 2024, pp. 17\,868--17\,879.

\bibitem{MPlib2023}
\BIBentryALTinterwordspacing
H.~S. Lab, ``Mplib: Motion planning library,'' 2023, accessed: 2024-09-28. [Online]. Available: \url{https://github.com/haosulab/MPlib}
\BIBentrySTDinterwordspacing

\end{thebibliography}
}

\clearpage
\appendix
\subsection{Real-World Experiment Details} \label{app:realworld}

\subsubsection{Dataset Collection, Pretraining and Training}
For the real-world experiments, we collected the LEAP Hand dataset and trained a model independently. We initially selected 78 daily objects from the YCB dataset~\cite{calli2017yale} and ContactDB~\cite{brahmbhatt2019contactdb}, then applied the DFC-based~\cite{dfc} grasp optimization method from~\cite{wang2022dexgraspnet} to generate 1,000 grasps per object, yielding a total of 78,000 grasps. Following a dataset filtering process, we obtained 24,656 grasps across 73 objects. The encoder network was first pretrained on the original dataset, and the entire model was then trained on the filtered dataset, as described in Sec.~\ref{sec:method}.

\subsubsection{Real-World Deployment Details}
We first scanned the objects listed in Tab.~\ref{tab:realworld} using AR Code~\cite{arcode2022}. After camera intrinsics~\cite{opencv_library} and extrinsics~\cite{chen2023easyhec} calibration, we estimated object poses using FoundationPose~\cite{wen2024foundationpose} and sampled point cloud uniformly on their surfaces. In this tabletop grasping setting, only top-down and side grasps are feasible, as other palm orientations would likely collide with the table. To address this, the model took as input the sampled object point clouds and a batch of LEAP Hand point clouds, which corresponded to 32 interpolated hand poses ranging from top-down to right-side orientations, enabled by our palm orientation control functionality. We randomly selected one of the top-5 grasps from the generated batch, ranked according to the same grasp energy calculation used during dataset generation~\cite{wang2022dexgraspnet}. We then use MPLib~\cite{MPlib2023} for arm motion planning to the desired end-effector pose. A PD controller is applied for grasp execution. 

\subsubsection{Experiment Result} 
\begin{figure*}[h]
    \centering
    \begin{subfigure}[b]{0.16\textwidth}
        \centering
        \includegraphics[width=\textwidth]{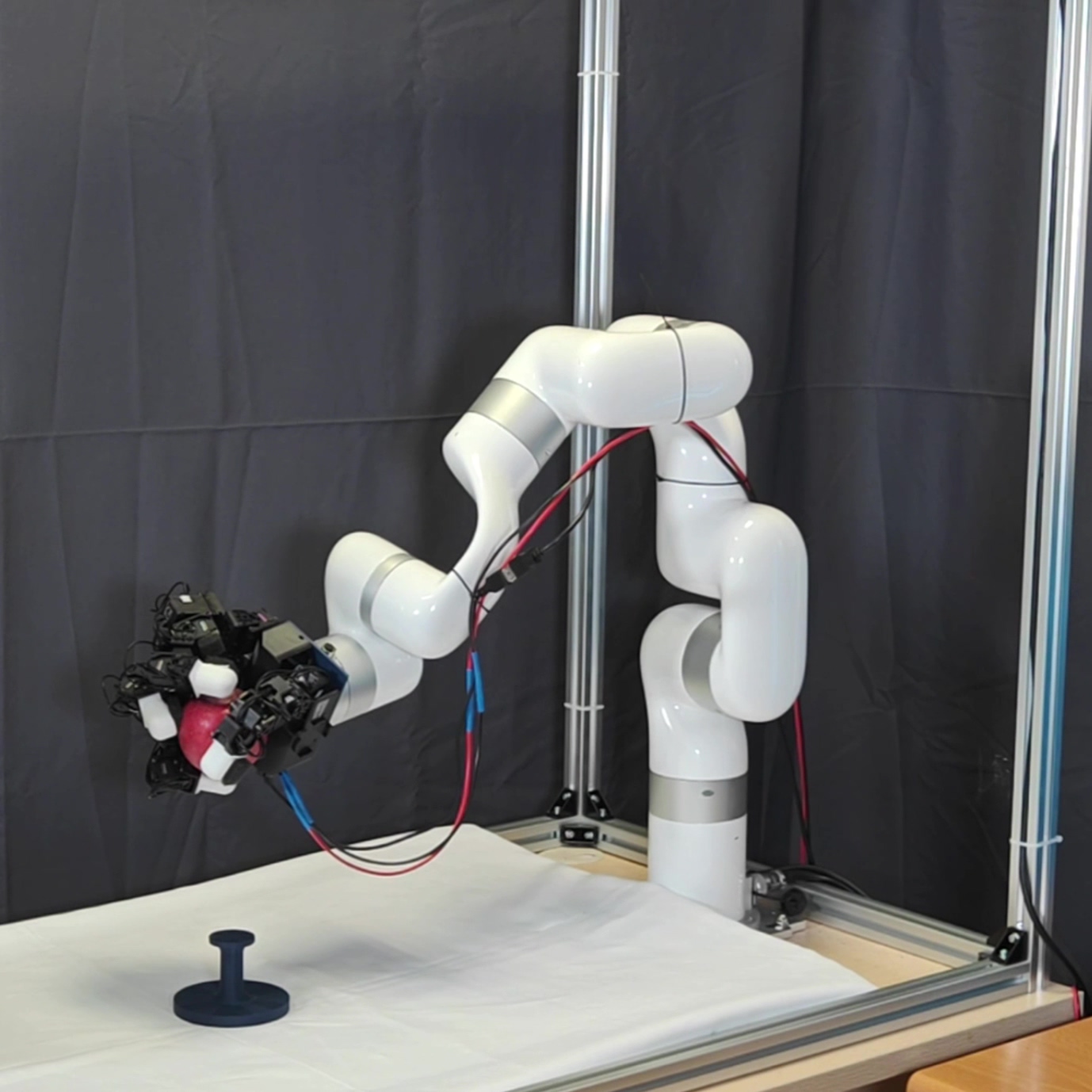}
        \caption{Apple}
    \end{subfigure}
    \begin{subfigure}[b]{0.16\textwidth}
        \centering
        \includegraphics[width=\textwidth]{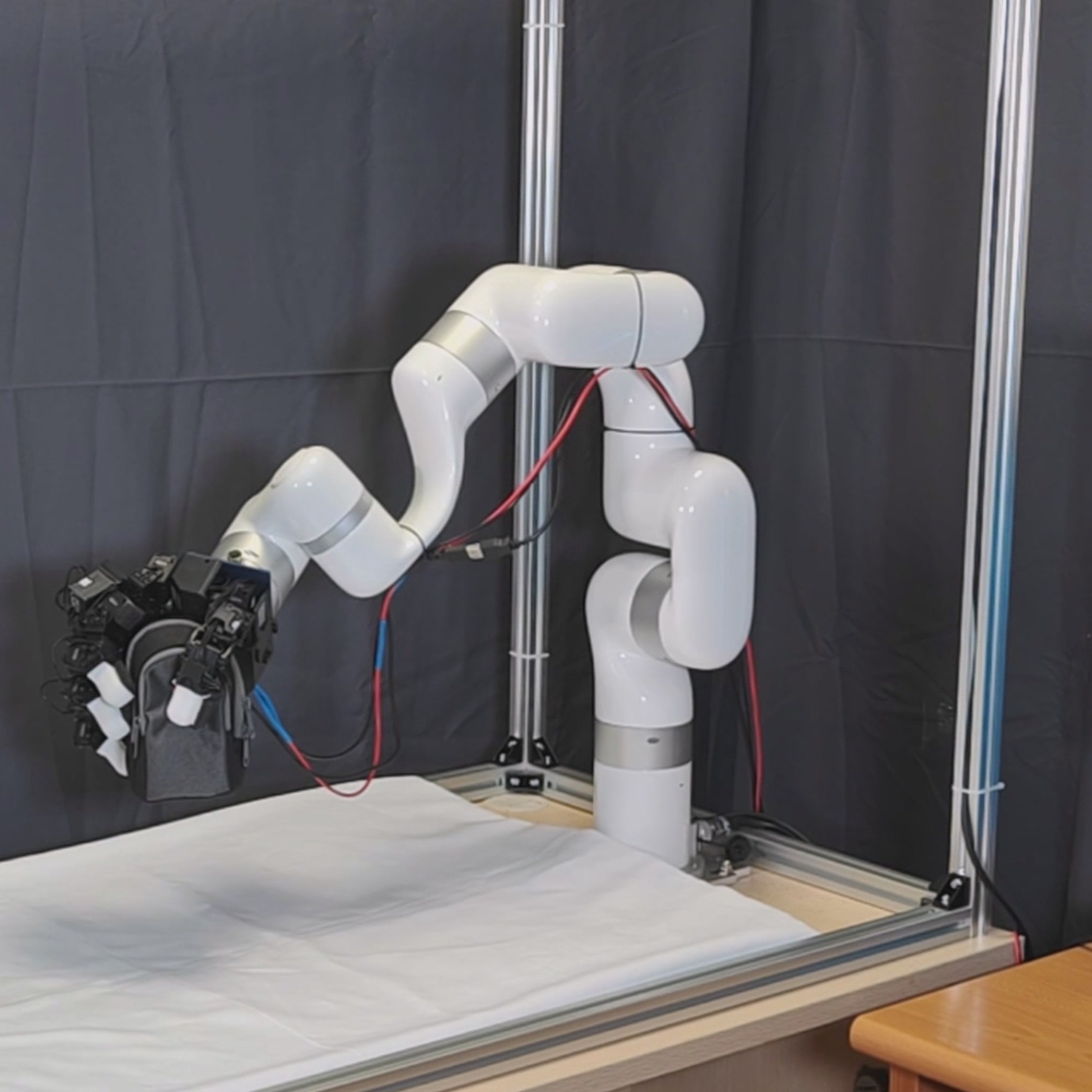}
        \caption{Bag}
    \end{subfigure}
    \begin{subfigure}[b]{0.16\textwidth}
        \centering
        \includegraphics[width=\textwidth]{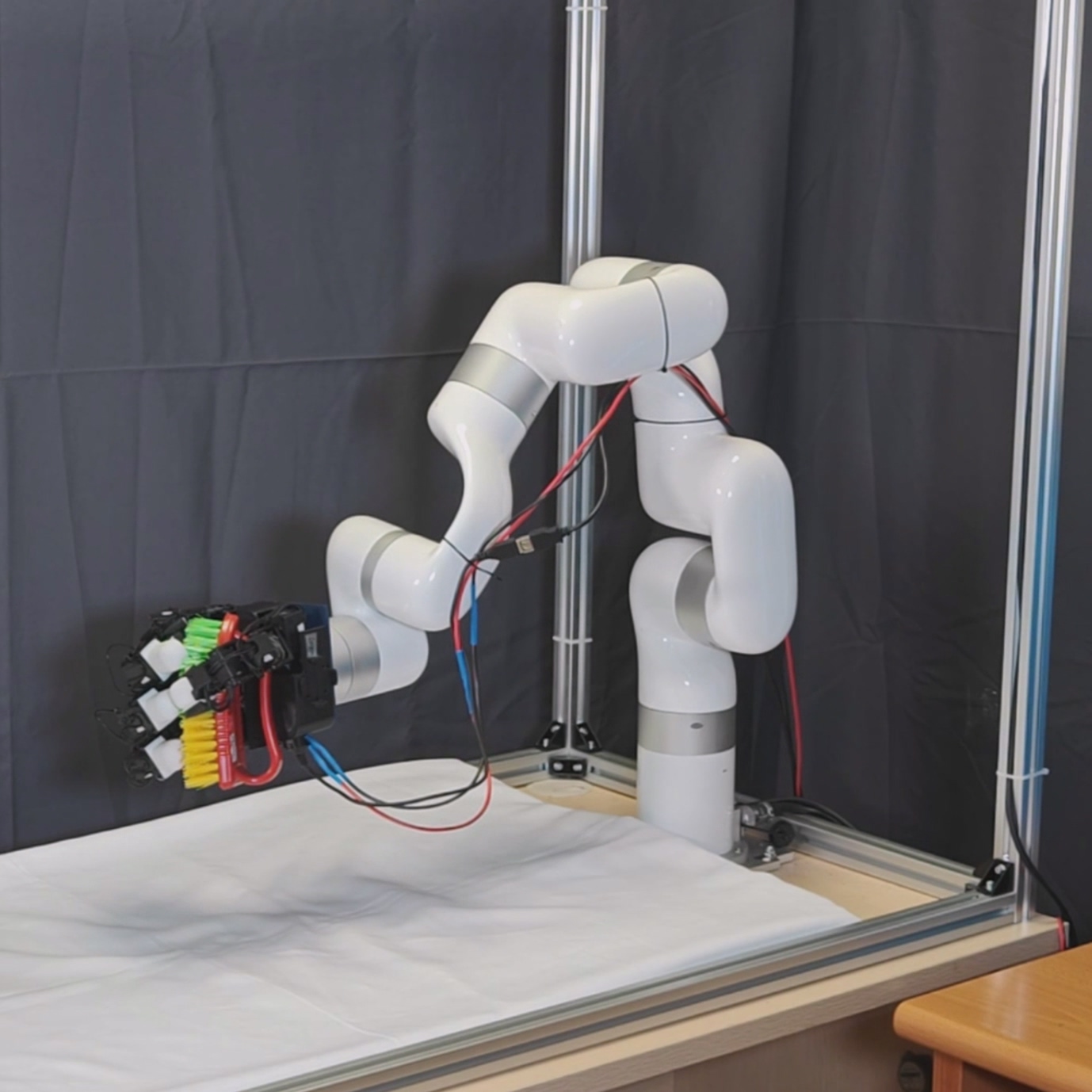}
        \caption{Brush}
    \end{subfigure}
    \begin{subfigure}[b]{0.16\textwidth}
        \centering
        \includegraphics[width=\textwidth]{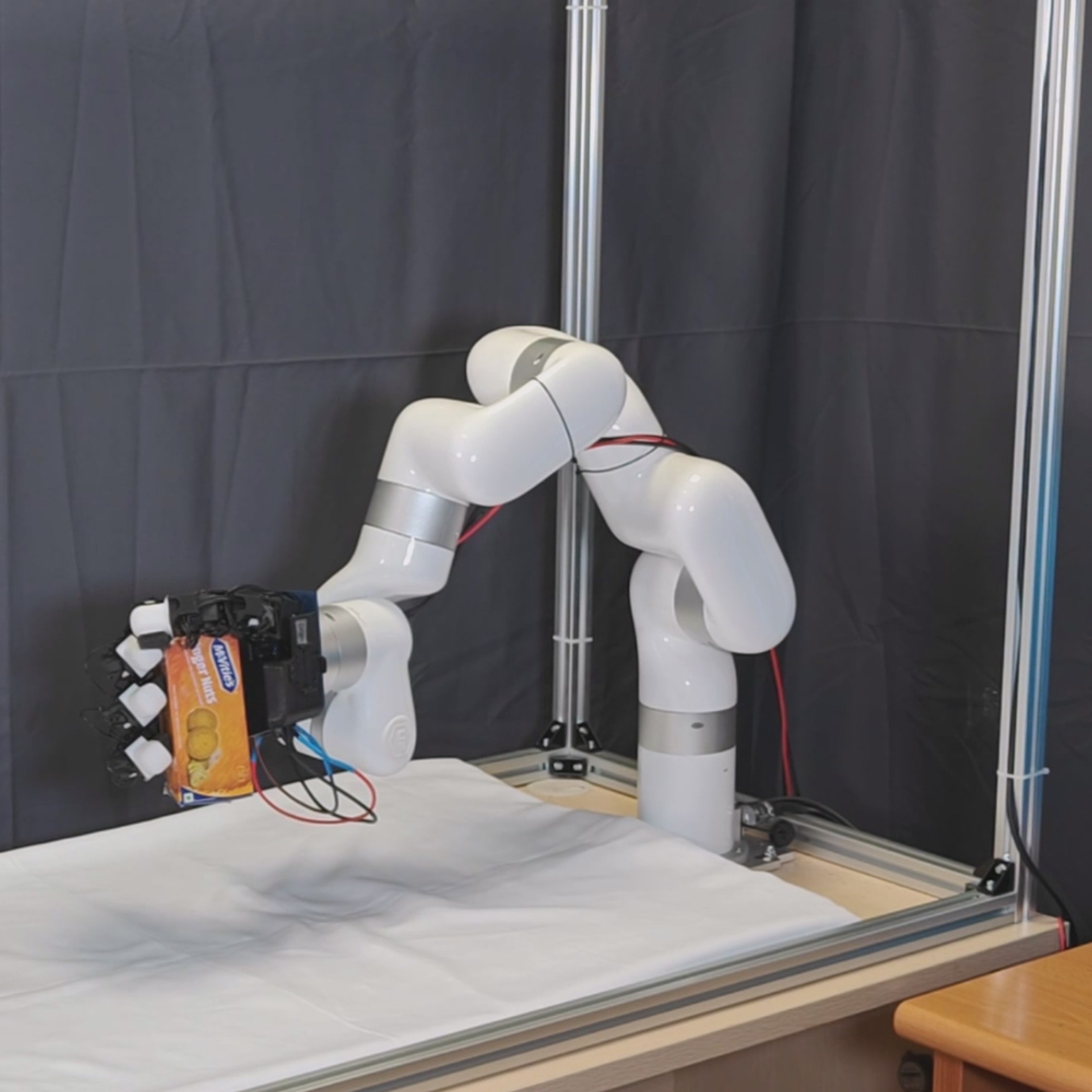}
        \caption{Cookie Box}
    \end{subfigure}
    \begin{subfigure}[b]{0.16\textwidth}
        \centering
        \includegraphics[width=\textwidth]{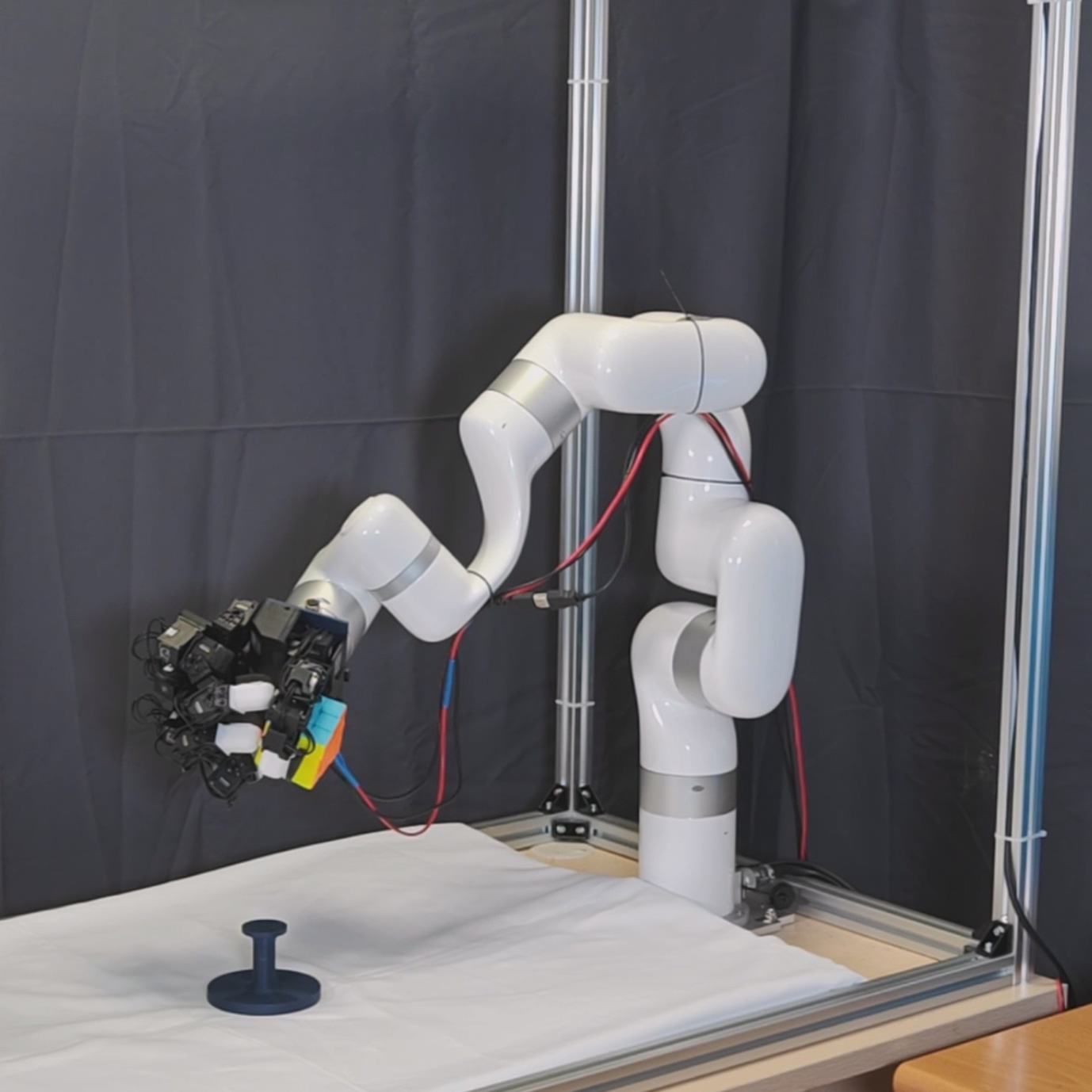}
        \caption{Cube}
    \end{subfigure}
    
    \vskip\baselineskip
    
    \begin{subfigure}[b]{0.16\textwidth}
        \centering
        \includegraphics[width=\textwidth]{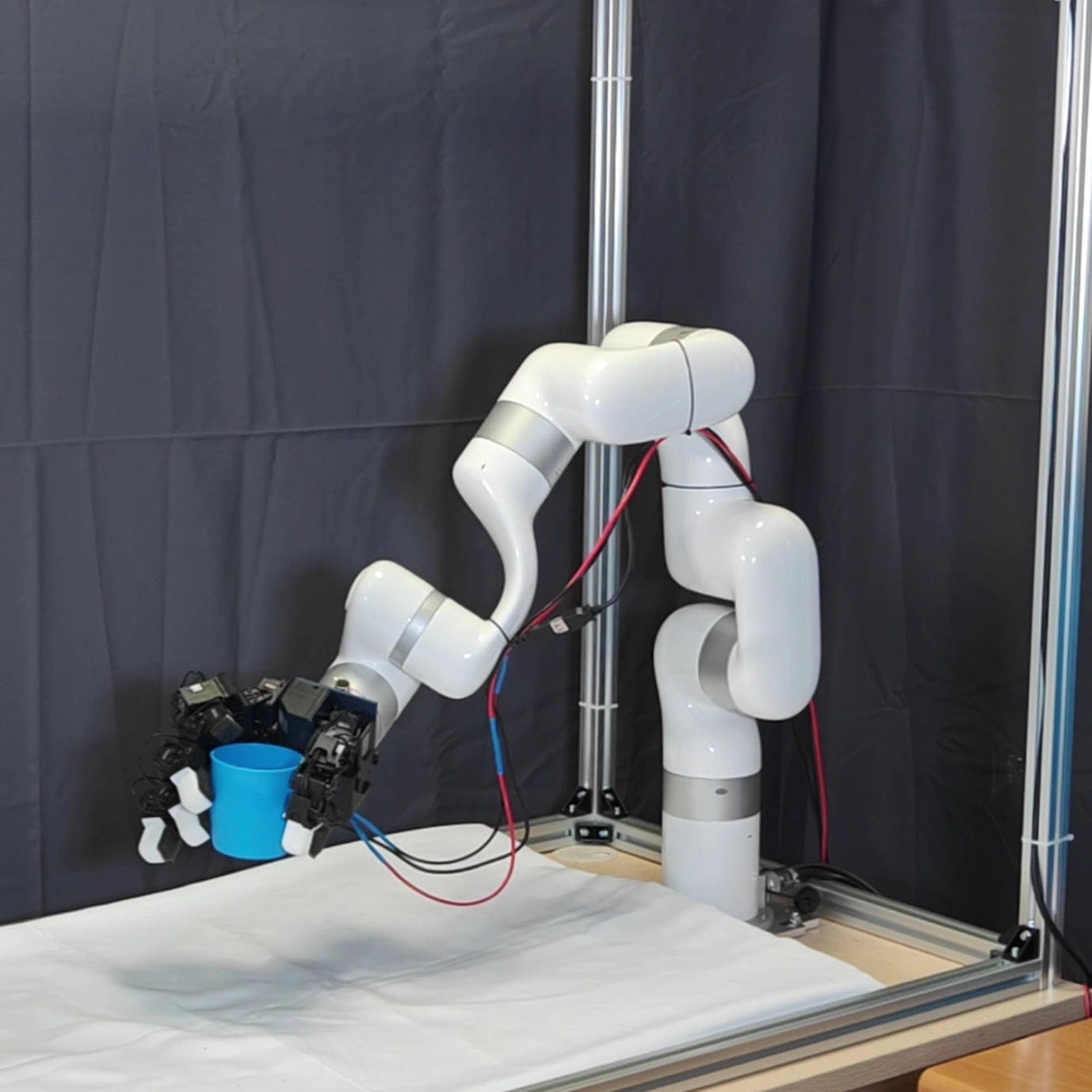}
        \caption{Cup}
    \end{subfigure}
    \begin{subfigure}[b]{0.16\textwidth}
        \centering
        \includegraphics[width=\textwidth]{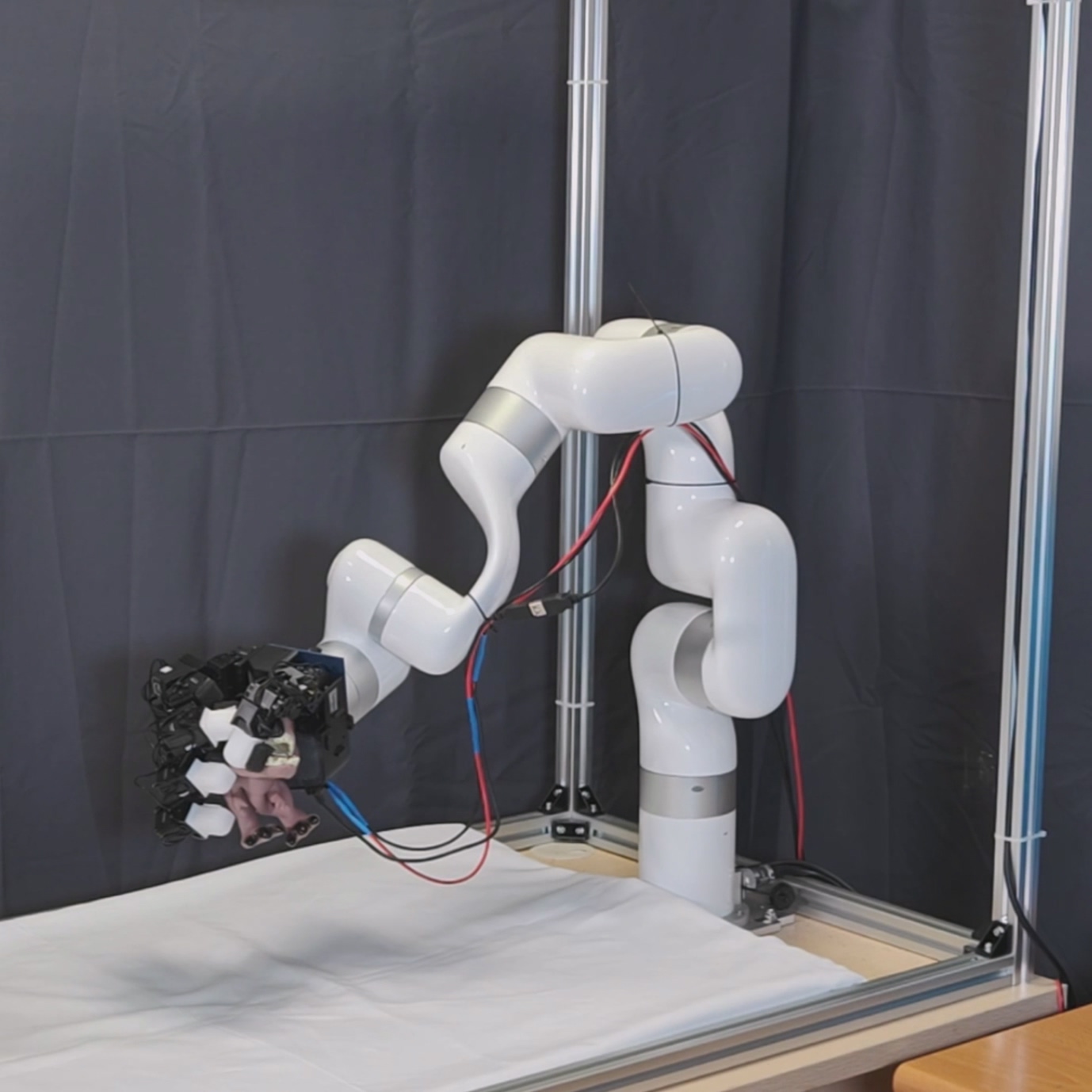}
        \caption{Dinosaur}
    \end{subfigure}
    \begin{subfigure}[b]{0.16\textwidth}
        \centering
        \includegraphics[width=\textwidth]{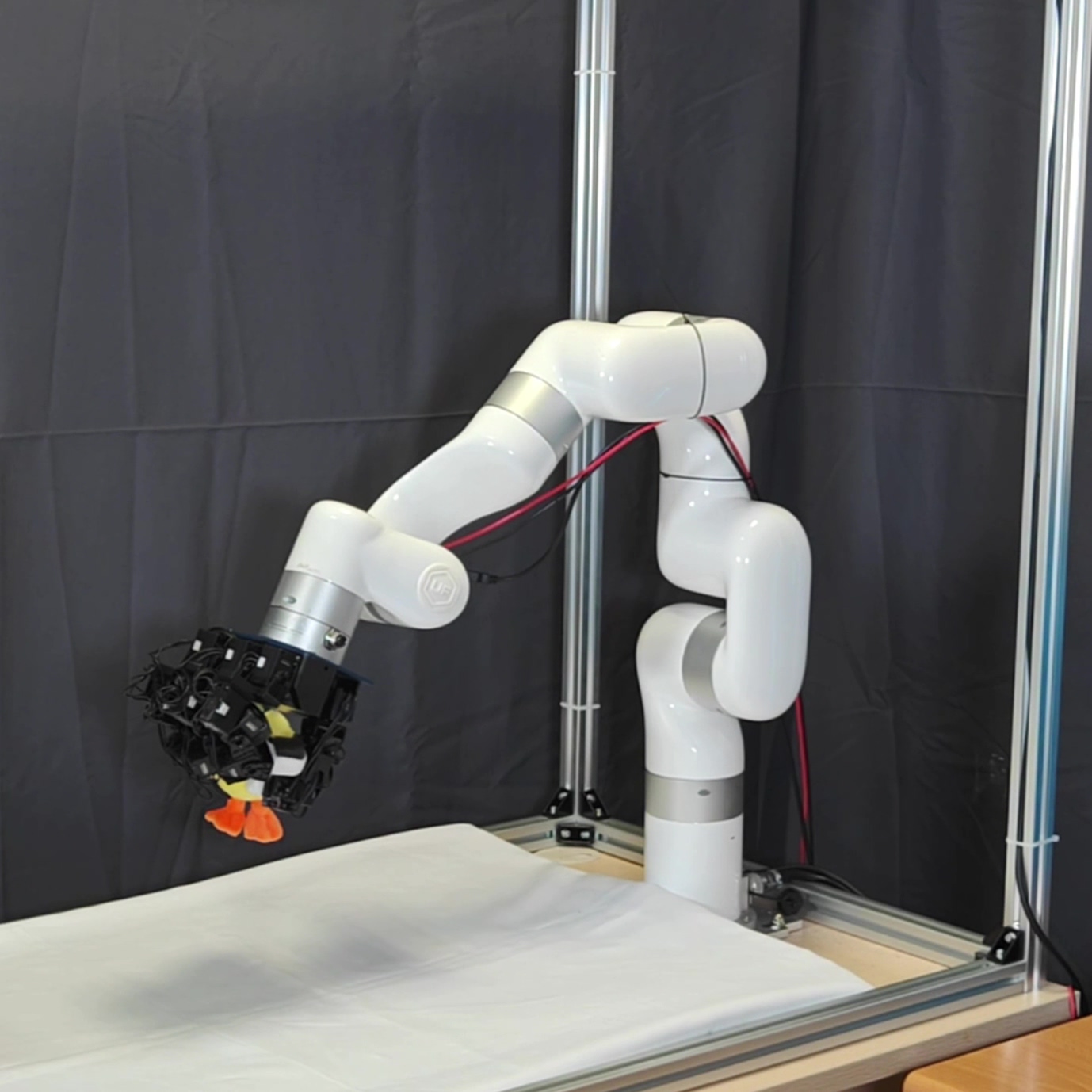}
        \caption{Duck}
    \end{subfigure}
    \begin{subfigure}[b]{0.16\textwidth}
        \centering
        \includegraphics[width=\textwidth]{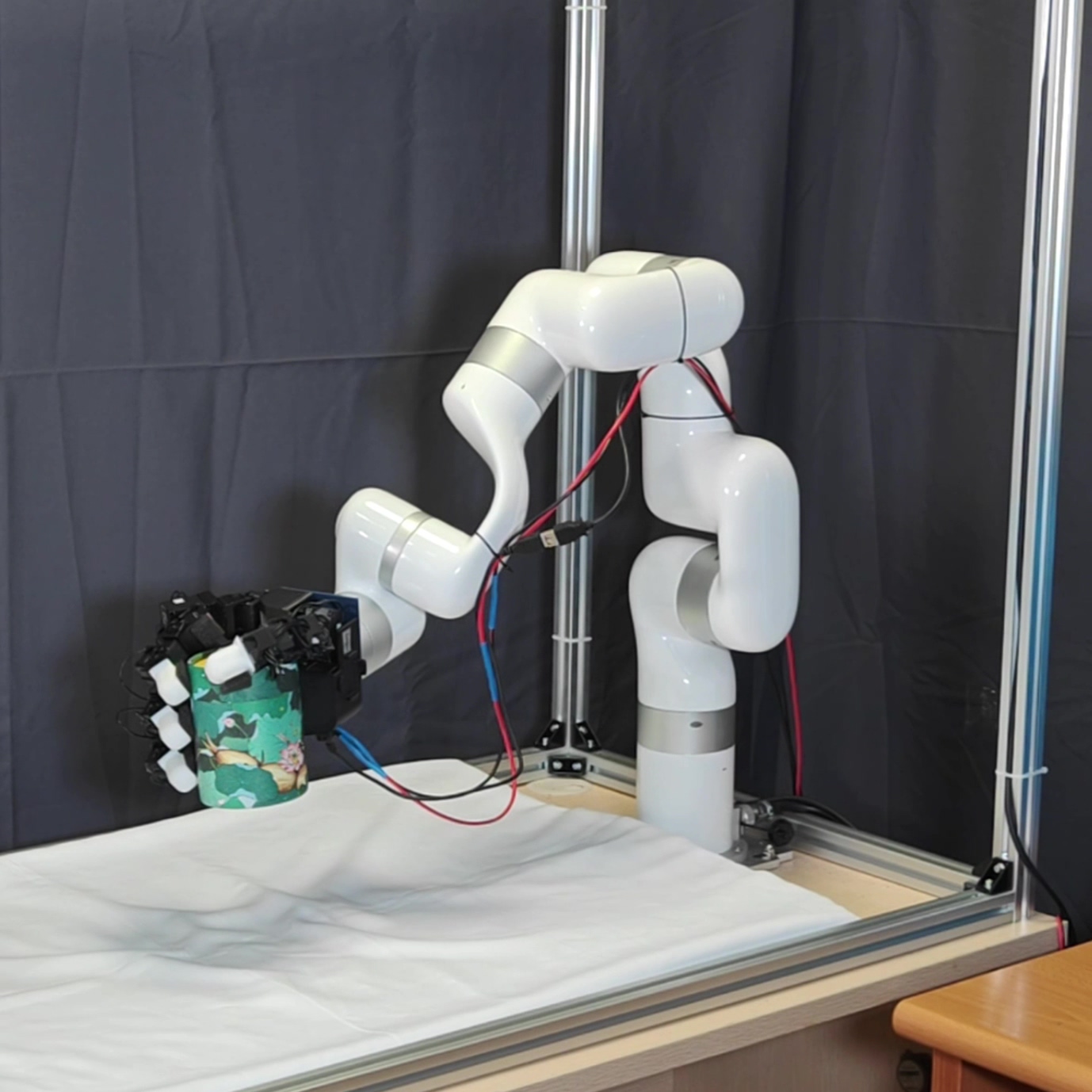}
        \caption{Tea Box}
    \end{subfigure}
    \begin{subfigure}[b]{0.16\textwidth}
        \centering
        \includegraphics[width=\textwidth]{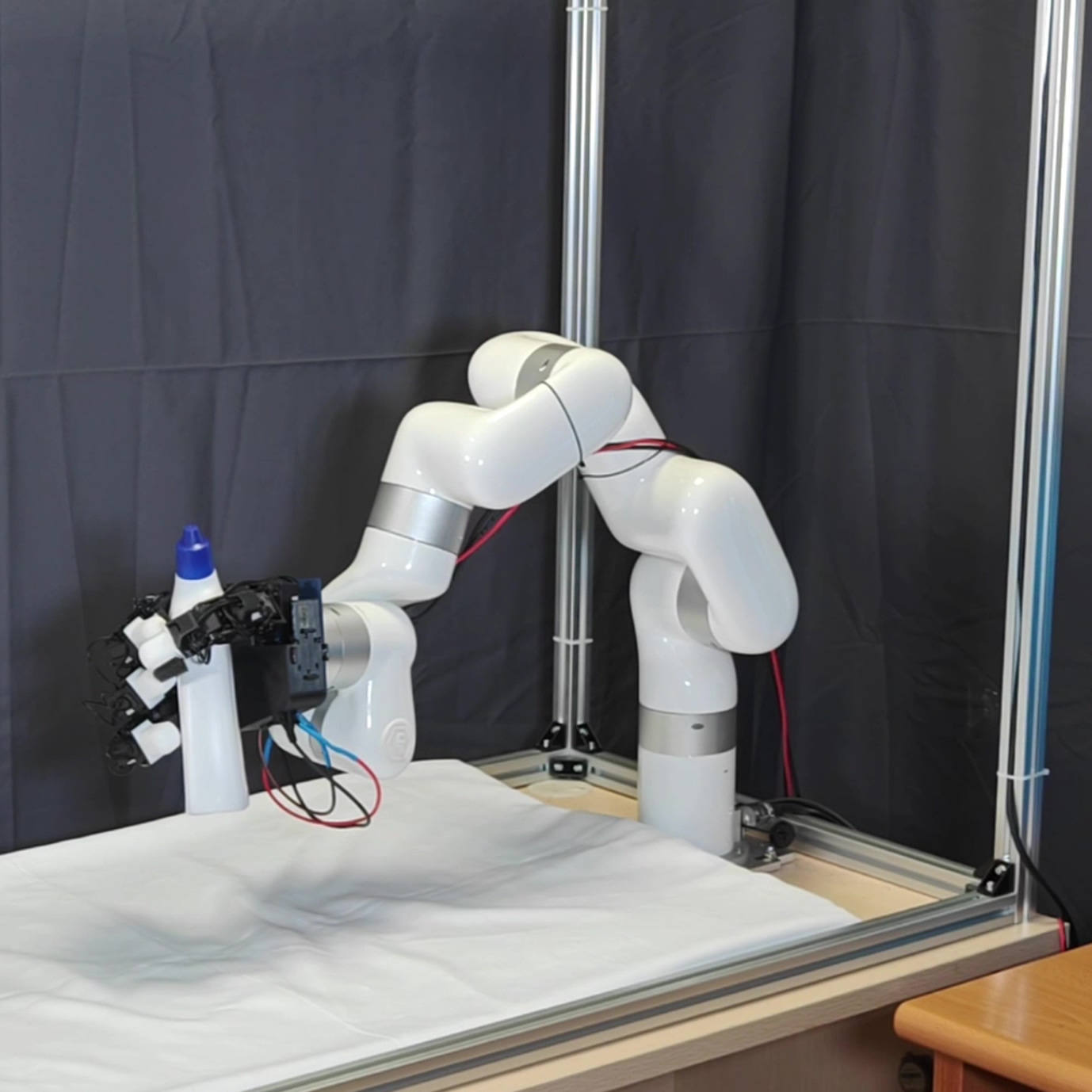}
        \caption{Toilet Cleaner}
    \end{subfigure}
    
    \caption{Real-world grasp demonstrations}
    \label{fig:realworld}
    \vspace{-10pt}
\end{figure*}

We tested 10 objects with various shapes, performing 10 grasping attempts for each object. The experimental results are shown in Tab.~\ref{tab:realworld} and Fig.~\ref{fig:realworld}. Our method achieved an average success rate of \textbf{89\%} across these 10 objects, demonstrating the effectiveness of our method in dexterous grasping and its generalizability to novel objects.

\begin{table}[ht]
    \centering
    \renewcommand\arraystretch{1.25}  
    \captionsetup{justification=centering, singlelinecheck=false}
    \resizebox{\linewidth}{!}{
        \begin{tabular}{ccccc}
            \toprule
            \textbf{Apple} & \textbf{Bag} & \textbf{Brush} & \textbf{Cookie Box} & \textbf{Cube}
            \\
            9/10 & 10/10 & 9/10 & 10/10 & 9/10
            \\
            \midrule
            \textbf{Cup} & \textbf{Dinosaur} & \textbf{Duck} & \textbf{Tea Box} & \textbf{Toilet Cleaner}
            \\
            7/10 & 9/10 & 8/10 & 8/10 & 10/10
            \\
            \bottomrule
        \end{tabular}
    }
    \caption{Real-world experiment results on unseen objects.}
    \label{tab:realworld}
    \vspace{-10pt}
\end{table}

\subsection{Zero-shot Generalization to Novel Hands Experiment} \label{app:zeroshot}
We trained the model separately on each of the three robotic hands and then validated it on the others without further training. As shown in Tab.~\ref{tab:zeroshot}, the results indicate that when transferring from high-DOF hands to low-DOF hands in a zero-shot setting, the model retains a certain level of performance. However, transferring in the opposite direction largely fails. We hypothesize that this difference arises because high-DOF hands have a much more complex configuration space, allowing the model to learn a broader range of articulation-invariant matching tasks, which can still perform well on the simpler articulation-invariant tasks required for low-DOF hands. In contrast, the configuration space of low-DOF hands is relatively simple, and when trained on these hands, the model can only master simple articulation-invariant matching tasks.

\begin{table}[htbp]
    \centering
    \renewcommand\arraystretch{1.25}  \captionsetup{justification=centering, singlelinecheck=false}
    \begin{tabular}{c|ccc}
        \toprule
        \multirow{2}{*} {\shortstack{\textbf{Training}\\ \textbf{Robot}}} 
        & \multicolumn{3}{c}{\textbf{Success Rate (\%) $\uparrow$}}
        \\
        \cline{2-4} 
        & Allegro & Barrett & ShadowHand
        \\ 
        \hline
        Allegro & (88.70) & 83.60 & 1.10
        \\
        Barrett & 42.40 & (84.80) & 6.90
        \\
        Shadowhand & 56.90 & 83.70 & (75.80)
        \\
        \bottomrule
    \end{tabular}
    \caption{Generalization results to novel hands.}
    \label{tab:zeroshot}
    \vspace{-15pt}
\end{table}

\begin{figure}[H] \centering
    \includegraphics[width=0.9 \linewidth]{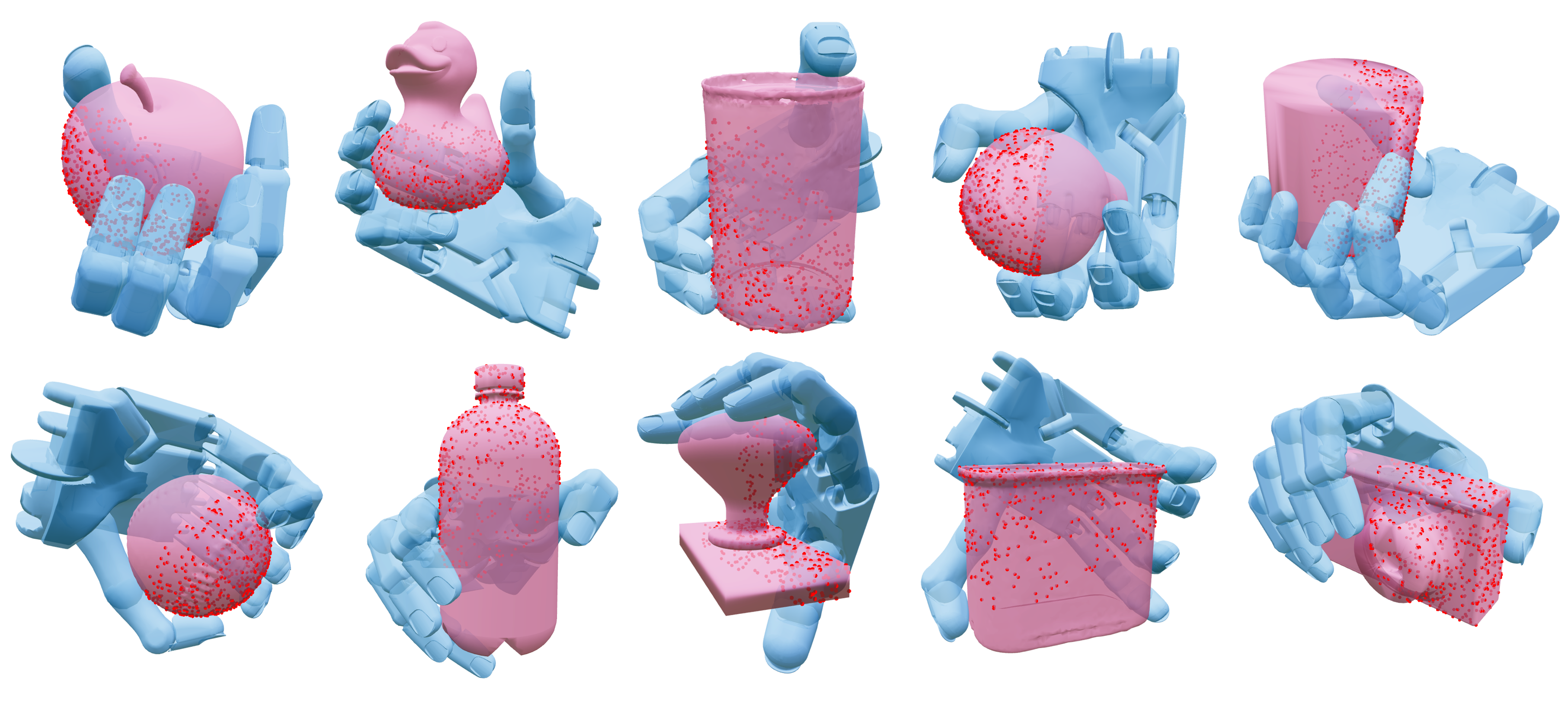}
    \vspace{-5pt}
    \caption{Grasp examples with partial object point clouds. Red points show the observed portion.}
    \label{fig:partial}
    \vspace{-5pt}
\end{figure}

\subsection{Partial Object Point Cloud Sampling} \label{app:partial}
Given the mesh of an object, we begin by randomly sampling $2 \times N_{\mathcal{O}}$ points. Next, a point is randomly sampled on a unit sphere, and the direction vector $r$ from this point to the origin is computed. For each point in the point cloud, we calculate the dot product between $r$ and the corresponding direction vectors $d_i$. We then remove half of the points with the smallest dot product values $r \cdot d_i$, leaving a subset of $N_{\mathcal{O}}$ points, which forms the partial object point cloud. This process is used to generate random point clouds during both training and evaluation.

\subsection{Grasp Controller} \label{app:controller}

\begin{figure}[H] \centering
    \vspace{-15pt}
    \includegraphics[width=0.9 \linewidth]{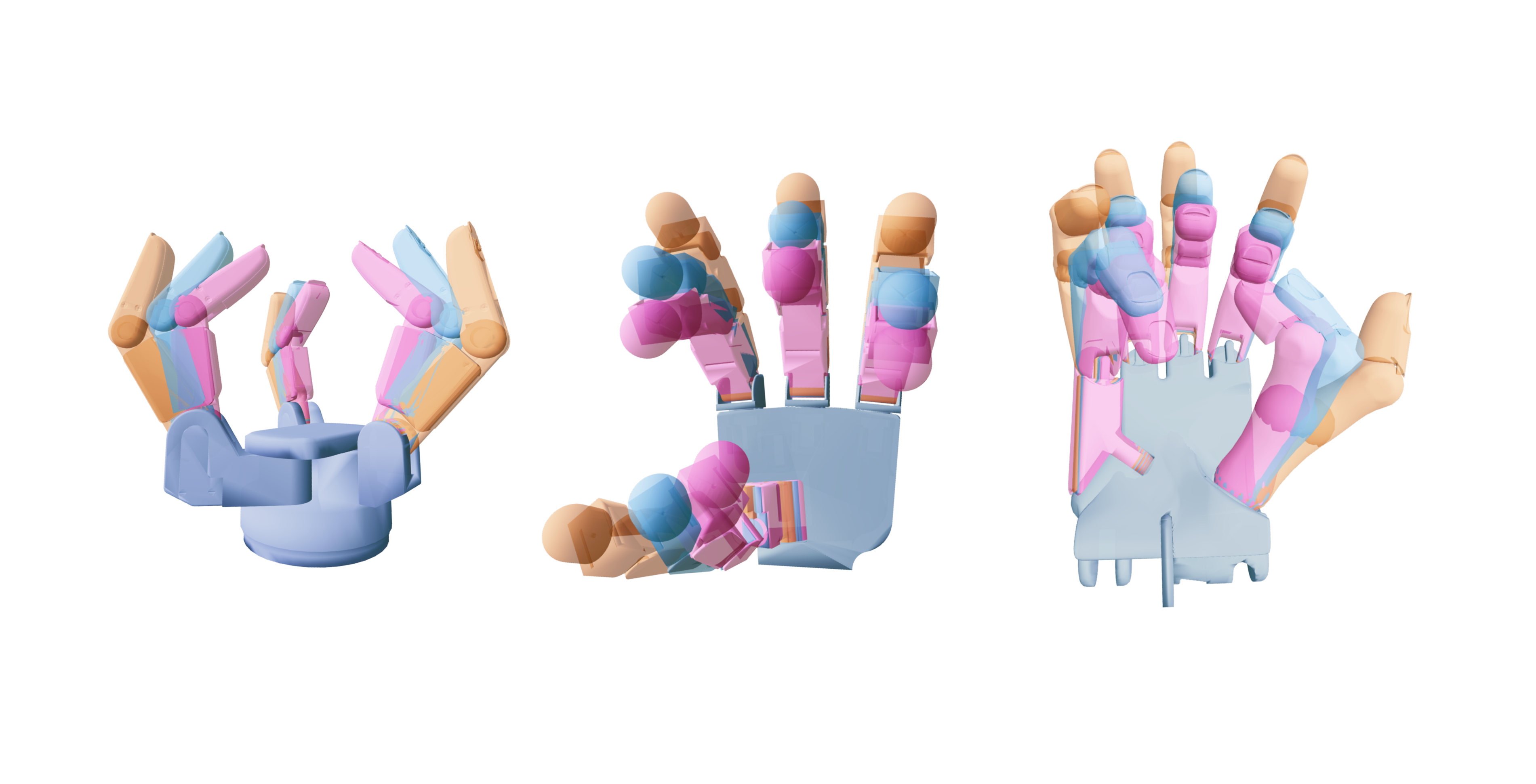}
    \vspace{-20pt}
    \caption{Visualization of the grasp controller's effect: blue indicates the predicted grasp pose, orange represents $q_{\text{outer}}$, and pink represents $q_{\text{inner}}$.}
    \label{fig:controller}
    \vspace{-5pt}
\end{figure}

To mitigate minor inaccuracies and subtle penetrations commonly found in generative methods, as well as the limitations of directly predicting a static grasp pose—which overlooks the forces exerted on contact surfaces—we developed a heuristic grasp controller to better simulate real-world grasping scenarios. The controller aims to generate a configuration $q_{\text{outer}}$ that is farther from the object's center of mass and a configuration $q_{\text{inner}}$ that is closer to the center of mass, based on the predicted pose. Fig.~\ref{fig:controller} illustrates the impact of the grasp controller.

\subsubsection{Evaluation Metric Details} \label{app:eval_detail}
In Isaac Gym, we evaluate the success of a grasp through a two-phase process. First, in the grasp phase, we use the previously described grasp controller to compute $q_{\text{outer}}$ and $q_{\text{inner}}$. We set the robot joint position to $q_{\text{outer}}$ with a position target at $q_{\text{inner}}$. Then we simulate for 1 second, equivalent to 100 simulation steps for the hand to close and grasp. In the second phase, we apply disturbance forces sequentially along six orthogonal directions, following the method in~\cite{li2023gendexgrasp}. These forces are defined as:
\begin{equation}
   F_{\pm xyz} = 0.5m/s^2 \times m_{\text{object}}
\end{equation}
where $m_{\text{object}}$ denotes the mass of the object. 

Our approach improves upon \cite{li2023gendexgrasp} by introducing a dynamic grasp phase, transitioning the evaluation from static to dynamic, and thereby significantly enhancing the rigor of the evaluation metric. In the original static validation, some grasps could hold objects in unstable positions. By introducing dynamic validation, these unstable grasps are less likely to succeed, resulting in a more stringent and accurate assessment of grasp quality. Moreover, static validation is prone to simulation errors, such as object penetration or robot self-collisions, which can incorrectly classify unstable grasps as successful. The dynamic method alleviates these issues, providing a more robust and reliable evaluation of grasp success. 

\begin{figure}[t] \centering
    \includegraphics[width=0.9 \linewidth]{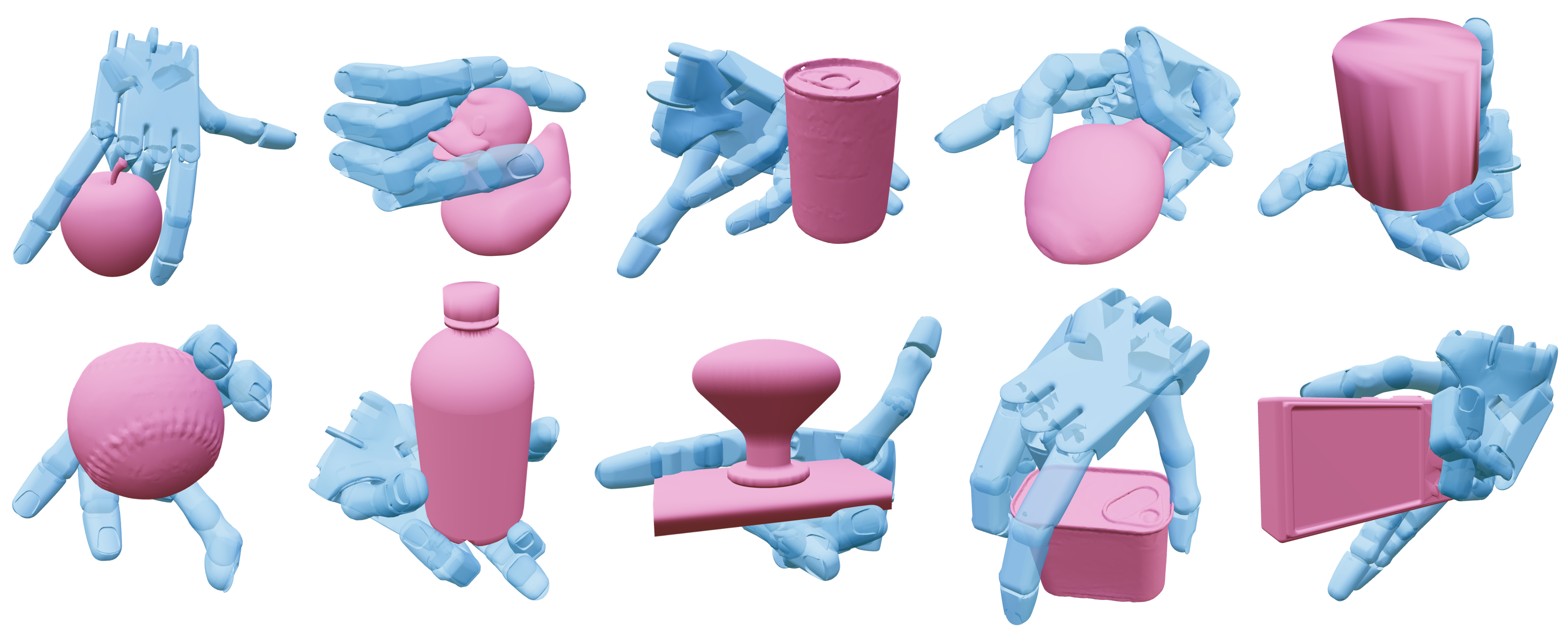}
    \vspace{-5pt}
    \caption{Grasp examples filtered out from the dataset that would otherwise be deemed successful under static metric.} \label{fig:static}
    \vspace{-15pt}
\end{figure}

Fig.~\ref{fig:static} illustrates several anomalous grasps that, despite appearing to fail, could still be judged as successful under the static metric. These grasps, either in an unstable state or exhibiting significant self-penetration, are impractical for real-world applications, highlighting the limitations of static validation.

\subsubsection{Dataset Filtering} \label{app:filter}
To address the suboptimal grasp quality, we applied a filtering process to the CMapDataset~\cite{li2023gendexgrasp}. Specifically, each grasp in the dataset was evaluated based on the success metrics defined in Sec.~\ref{sec:evalmetric} and Appendix~\ref{app:eval_detail}. We then store the relative 6D pose and joint values of every successful grasp in the filtered dataset.

\subsection{Baseline Description}

\subsubsection{DFC~\cite{dfc}}
Since DFC is a purely optimization-based method, the speed of generating grasps is particularly slow. Therefore, we evaluate it using the original CMapDataset, which was primarily generated by the DFC method. As the dataset generation process also minimizes the hand prior energy and penetration energy described in \cite{li2023gendexgrasp}, and some generated grasps may have already been filtered, the evaluation results are likely better than DFC's actual performance.

\subsubsection{GenDexGrasp~\cite{li2023gendexgrasp}}
We used the filtered grasp dataset to train the model, where the contact heatmap was generated using the aligned distance as described in the paper. The GenDexGrasp model was trained with default hyperparameters. In Tab.~\ref{tab:gendexgrasp}, we compared the results of the open-source pretrained model with those of our trained model, demonstrating that our filtered dataset is of higher quality.

\begin{table}[htbp]
    \centering
    \renewcommand\arraystretch{1.25}  
    \captionsetup{justification=centering, singlelinecheck=false}
    \begin{tabular}{c|ccc|c}
        \toprule
        \multirow{2}{*} {\textbf{Method}} 
        & \multicolumn{4}{c}{\textbf{Success Rate (\%) $\uparrow$}}
        \\ 
        \cline{2-5} 
        & Allegro & Barrett & ShadowHand & Avg.
        \\ \hline
        pretrain & 51.00 & 63.80 & 44.50 & 53.10
        \\
        train & 51.00 & 67.00 & 54.20 & 57.40
        \\
        \bottomrule
    \end{tabular}
    \caption{GenDexGrasp Result Comparision}
    \vspace{-5px}
    \label{tab:gendexgrasp}
\end{table}

\subsubsection{ManiFM~\cite{xu2024manifoundation}}
Due to the unavailability of pretrained models for Barrett and ShadowHand, our evaluation was restricted to the Allegro pretrained model. Considering the fundamental differences between point-contact and surface-contact grasps, we optimized the controller's hyperparameters for improved performance of ManiFM. Nevertheless, despite the seemingly high quality of the generated grasps, the inherent instability of point-contact grasps posed significant challenges in achieving a high success rate during simulation.

\subsubsection{GeoMatch~\cite{attarian2023geometry}}
Although GeoMatch is a keypoint matching-based method that supports cross-embodiment and shares similarities with our approach, we faced challenges in reproducing its results due to the absence of pretrained models and insufficient details regarding the data file formats, which remained unsolved in its repository's issues as well. Consequently, it was not included as a baseline for comparison.

\subsection{Network Architecture} \label{app:arch}

\subsubsection{Point Cloud Encoder} \label{app:arch_encoder}
To map robot and object features into a shared feature space, enhancing the network's ability to learn correspondences between them, we employed identical architectures for both the robot and object encoders. Our encoder design is based on the DGCNN~\cite{wang2019dynamic} architecture, as implemented in~\cite{eisner2024deep}. Notably, this implementation omits the original layer-wise re-computation of K-nearest neighbors (KNN) for graph construction, resulting in a ``Static Graph CNN''. In our setup, K is set to 32, meaning that each point's receptive field is much smaller than the total number of points in the cloud ($N_{\mathcal{R}}=512$). This constraint limits the ability of the per-point feature extraction process to capture global information, which poses a challenge for the object encoder, as it struggles to learn comprehensive geometric shape features.

We experimented with the original dynamic graph structure, but it led to a decline in pretraining performance. We hypothesize that, for configuration-invariant learning objectives, local structural information in the point cloud is critical, and the network needs to be reinforced to capture this. The dynamic graph structure tends to learn similar structures across different fingers, which, while beneficial for segmentation tasks, is less suited to our specific learning goals. The impact of varying network architectures and feature learning strategies will be further explored in future work.

\begin{figure}[t] \centering
    \includegraphics[width=0.9 \linewidth]{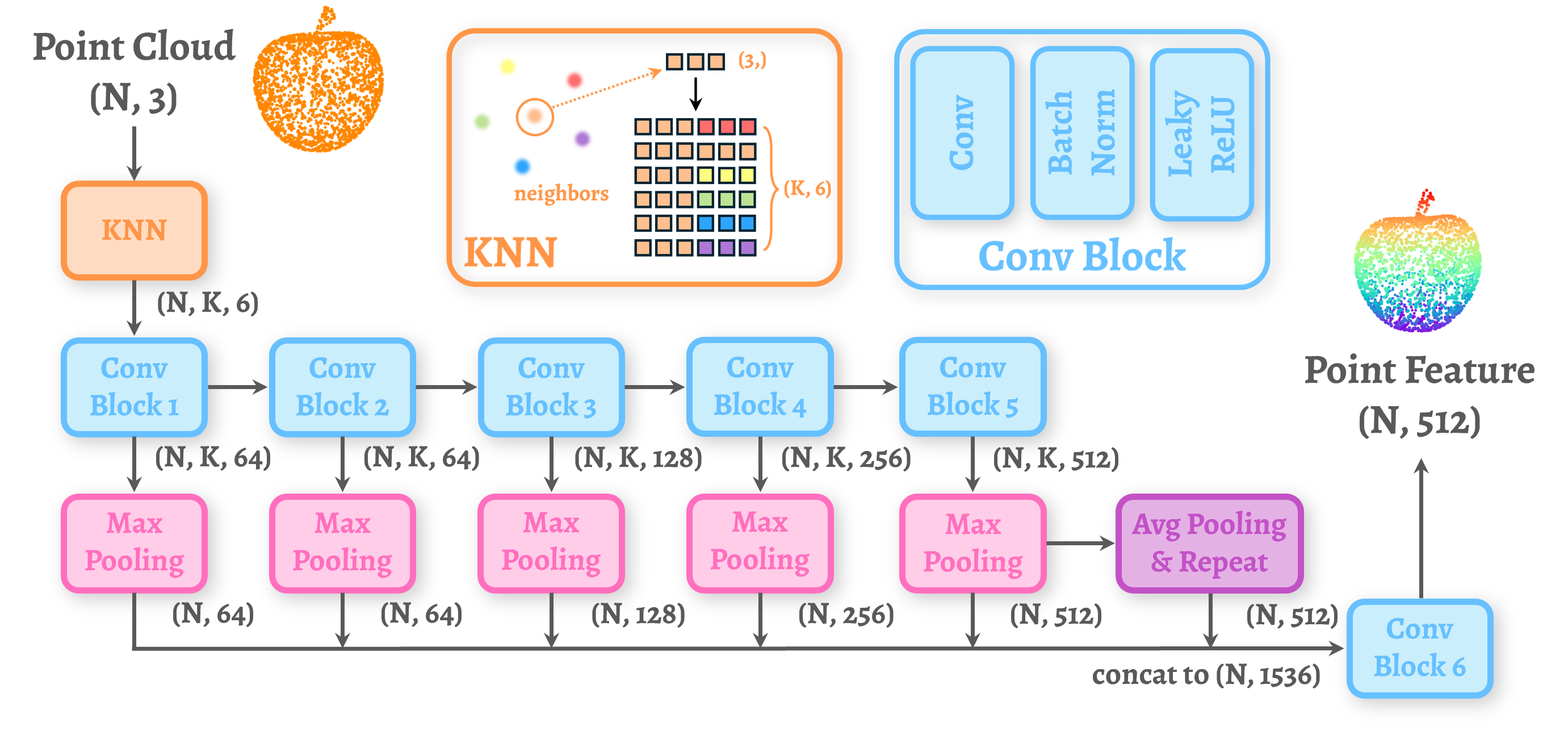}
    \vspace{-5pt}
    \caption{Point cloud encoder architecture.} \label{fig:encoder}
    \vspace{-15pt}
\end{figure}

Consequently, our encoder follows a ``Static Graph CNN'' architecture with five convolutional layers. After the last convolution, a global average pooling layer generates a global feature concatenated with features from all previous layers. This combined output is passed through a final convolutional layer, projecting into the embedding dimension. The architecture is illustrated in Fig.~\ref{fig:encoder}, where the LeakyReLU activation function uses a negative slope of 0.2.

\subsubsection{Cross-Attention Transformer} \label{app:arch_tf}
We followed the architectural design from~\cite{eisner2024deep}, utilizing a multi-head attention block with 4 heads. The implementation details can be found in the code.

\subsubsection{Kernel MLP} \label{app:arch_mlp}
We adopted the same hyperparameters design as \cite{eisner2024deep}. Specifically, the MLP consists of two hidden layers with feature dimensions of 300 and 100, respectively, along with the ReLU activation function.

\subsection{Dataset Preprocessing}

\subsubsection{URDF File Preprocessing}
To facilitate optimization, we introduce six virtual joints between the world frame and the robot's root link: three prismatic joints representing translation $(x, y, z)$ and three revolute joints representing rotation $(roll, pitch, yaw)$. These virtual joints are incorporated into the robot's URDF file and treated equivalently to other joints to simplify the computation of the Jacobian matrix. Furthermore, virtual links are added to the distal ends of each tip link to address potential errors in the 6D pose during optimization, ensuring consistent constraints across all links despite reduced rotational restrictions.

\subsubsection{Robot Point Cloud Sampling} \label{subsubsec:robot_pc_sample}
To extract the stored point clouds $\left\{\mathbf{P}_{\ell_i}\right\}_{i=1}^{N_{\ell}}$ from the URDF file of a specific robot, we first sample 512 points from the mesh of each link. We then apply the Farthest Point Sampling (FPS) algorithm to the complete point cloud, selecting 512 points, denoted as $N_R$ in our method. These point clouds are stored separately for each distinct link.

This process guarantees that, for any joint configuration, our point cloud forward kinematics model, $\text{FK}\left(q, \left\{\mathbf{P}_{\ell_i}\right\}_{i=1}^{N_{\ell}}\right)$, can map joint configurations to corresponding point clouds at new poses. This ensures consistent point cloud correspondence across different poses, a key advantage for our pretraining methodology.

\subsubsection{Object Point Cloud Sampling}
Starting with the mesh file of an object, we initially sample 65,536 points. For each training iteration, we randomly select 512 points from this set and apply Gaussian noise $\mathcal{N}(0, 0.002)$ for data augmentation. This strategy improves the model's generalization across different object shapes.

\subsection{Matrix Block Computation}
To address the high GPU memory demands of using the MLP kernel function to compute $\mathcal{D(R,O)}$, we implemented a matrix block computation strategy to optimize memory usage. After experimentation, we ultimately chose to divide the entire matrix into $4\times 4$ blocks for computation, which reduces memory consumption by approximately 34\% while maintaining similar computation time.

\end{document}